\pdfoutput=1

\documentclass[11pt]{article}

\usepackage[preprint]{acl}

\usepackage{booktabs}

\usepackage{times}
\usepackage{latexsym}

\usepackage[T1]{fontenc}

\usepackage[utf8]{inputenc}

\usepackage{microtype}

\usepackage{inconsolata}

\usepackage{graphicx}

\usepackage{amsmath}

\title{EduVidQA: Generating and Evaluating Long-form Answers to Student Questions based on Lecture Videos}

 \author{Sourjyadip Ray\textsuperscript{1}, Shubham Sharma\textsuperscript{2}, Somak Aditya\textsuperscript{1*}, Pawan Goyal\textsuperscript{1*} \\
  \texttt{\{sourjyadipray@kgpian, saditya@cse, pawang@cse\}.iitkgp.ac.in}
  \\ 
  \texttt{amnour.rajsubham@gmail.com}
 \\
         \textsuperscript{1}Indian Institute of Technology, Kharagpur \\ 
         \textsuperscript{2}Panjab University, Chandigarh}

\begin{document}
\maketitle
\begin{abstract}
As digital platforms redefine educational paradigms, ensuring interactivity remains vital for effective learning. This paper explores using Multimodal Large Language Models (MLLMs) to automatically respond to student questions from online lectures - a novel question answering task of real world significance. We introduce the EduVidQA Dataset with 5252 question-answer pairs (both synthetic and real-world) from 296 computer science videos covering diverse topics and difficulty levels. To understand the needs of the dataset and task evaluation, we empirically study the qualitative preferences of students, which we provide as an important contribution to this line of work. Our benchmarking experiments consist of 6 state-of-the-art MLLMs, through which we study the effectiveness of our synthetic data for finetuning, as well as showing the challenging nature of the task. We evaluate the models using both text-based and qualitative metrics, thus showing a nuanced perspective of the models' performance, which is paramount to future work. This work not only sets a benchmark for this important problem, but also opens exciting avenues for future research in the field of Natural Language Processing for Education.
\end{abstract}
\let\thefootnote\relax\footnotetext{* indicates equal supervision}
\let\thefootnote\relax\footnotetext{Code and data: \url{https://github.com/sourjyadip/eduvidqa-emnlp25}}
\section{Introduction}

\begin{quote}
“Tell me and I forget. Teach me and I remember. Involve me and I learn.”  \\
\emph{\href{https://www.degruyter.com/document/doi/10.1515/9781400852550/html}{- Xunzi (English Translation), 818 AD} }
\end{quote}

This enduring insight, attributed to the ancient Chinese philosopher Xun Kuang (Xunzi), eloquently captures the essence of interactive learning. Indeed, the importance of active learner participation has been extensively examined in educational and pedagogical research \cite{barker1994designing, reeves1997effective, mcintyre1998experiment, pradono2013method}, particularly in the age of web-based instruction. Influential frameworks such as the ICAP model \cite{chi2014icap} demonstrate how deeper cognitive engagement directly fosters more effective learning. Consequently, \textit{“active learning”} is often defined as requiring students to engage cognitively and meaningfully with the material \cite{bonwell1991active}—to analyze, synthesize, and evaluate rather than passively absorb information \cite{king1993sage}. One widely recognized strategy for cultivating such engagement is the encouragement of student-generated questions during classes or lectures \cite{graesser1994question}, underscoring the crucial role that curiosity and interaction play in successful education. 

\begin{figure}[!t]
    \centering
    \includegraphics[width=0.8\columnwidth]{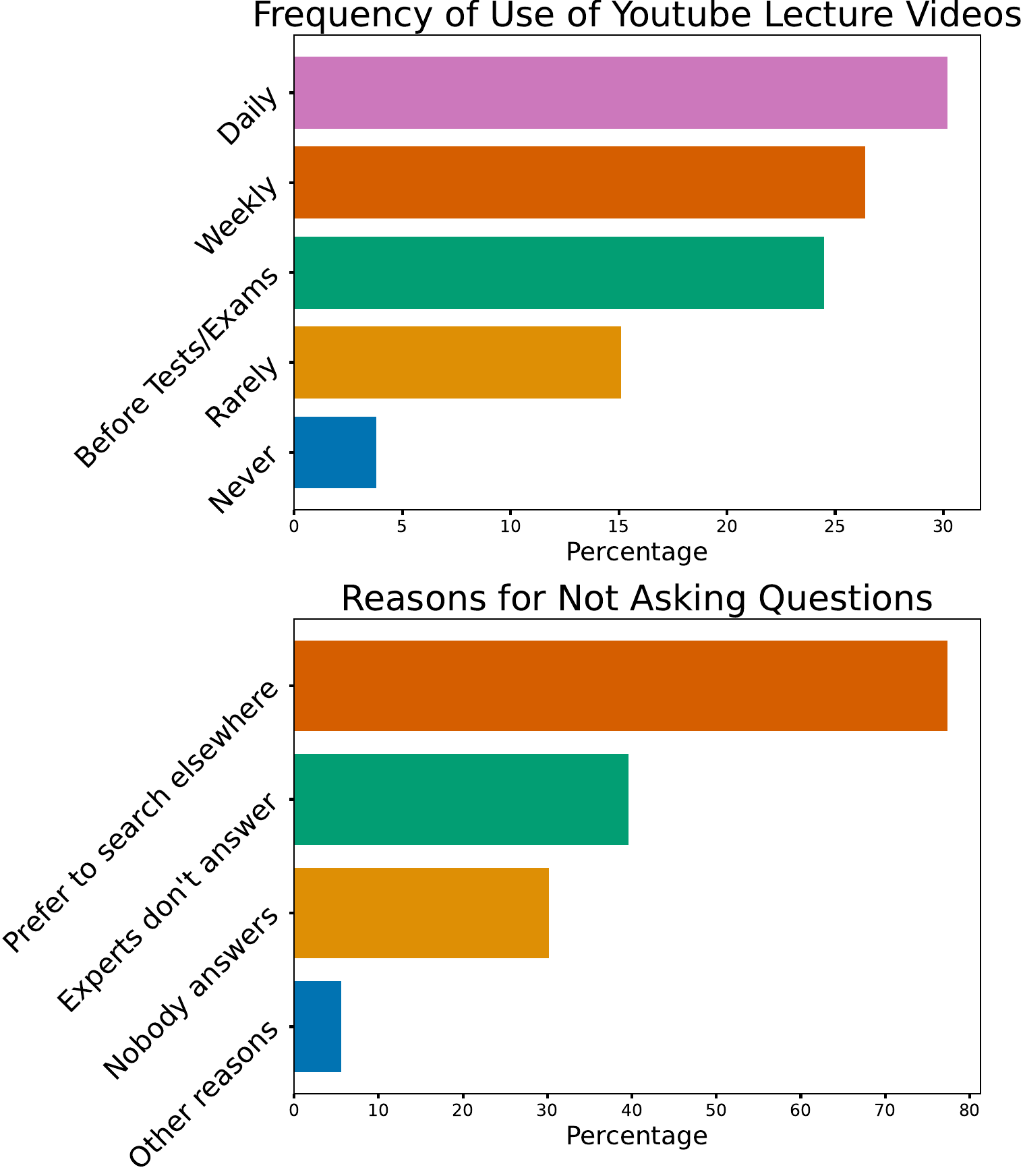}
    \caption{Results from the preliminary study on student interaction with online lecture videos showing i) frequency of use of online lecture videos to learn new topics and ii) reasons for not asking questions in the comments section}
    \label{fig:study}
\end{figure}

\begin{figure*}
    \centering
    \includegraphics[width=0.85\textwidth]{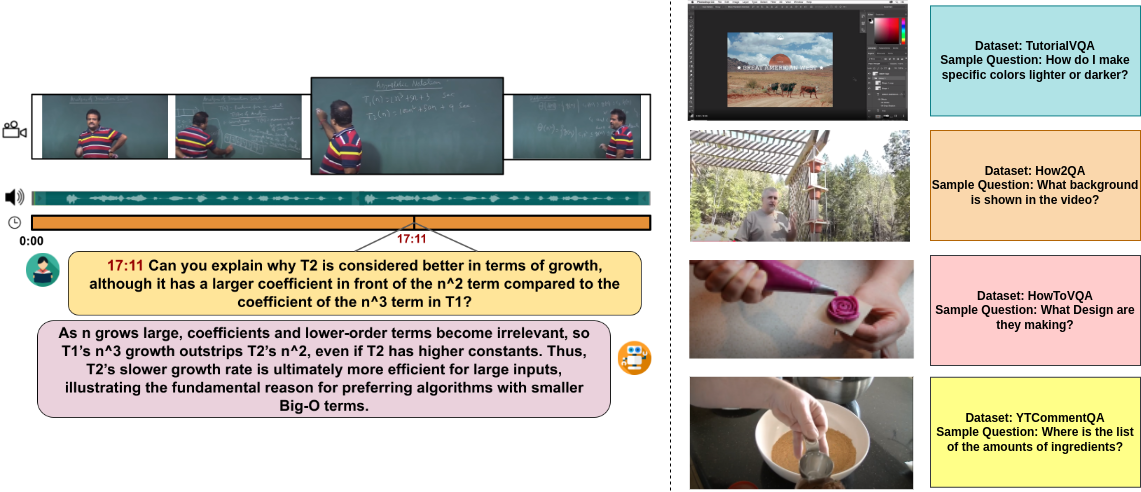}
    \caption{\textit{\textbf{Left:}} Example data point from the EduVidQA dataset. \textit{\textbf{Right:}} Sample frames and questions from related video datasets, TutorialVQA \cite{colas2019tutorialvqa}, How2QA \cite{li2020hero}, HowToVQA \cite{yang2021just} and YTCommentQA \cite{yang2024ytcommentqa}.}
    \label{fig:data_example}
\end{figure*}

With the ever-growing popularity of Youtube as a content sharing platform, institutions and individuals have been posting educational videos online to democratize and distribute learning resources throughout the world. However, a major disadvantage of learning from online lectures has been the lack of interaction with educators \cite{stecula2022advantages}. \\
\textbf{Motivation Study:} We do an initial exploration of this problem by conducting a survey with 52 undergraduate (75.5\%) and graduate students (24.5\%) enrolled in Science, Technology, Engineering, and Mathematics (STEM) education programs. 56.8\% of these students were using online lecture videos to study at least once a week, while 24.5\% of the students only used these videos before tests or exams. A majority of students (67.9\%) have felt the need to ask questions about the content while watching educational videos, indicating a strong interest in interactive learning. However, despite the desire to ask questions, only 17\% of the students actually asked content related questions in the comments section. Figure \ref{fig:study} shows the reasons for the students not asking questions in the comments section, as well as the satisfaction with the responses they received when asking such questions. This study highlights the need for a dependable, expert-level question answering system. The granular level details of the study are discussed in the Appendix A.  \\
With the growing impressive capabilities of Multimodal Large Language Models (MLLMs) in various different tasks \cite{wang2024comprehensive}, and the popularity of using synthetic data for LLM finetuning \cite{chen2024diversity, li2023synthetic} we explore the following 2 research questions: \textbf{RQ1: \textit{How well can MLLMs automatically generate answers to questions based on online lectures, asked by students?}} and \textbf{RQ2: \textit{Does finetuning with synthetic data help with this task?}}      
The key contributions of this work are the following: \\
\textbf{C1 (\textit{Dataset})}: We first collect 270 real world questions from Youtube videos and answers from domain experts, and observe poor performance of MLLMs on this data. To tackle this problem further, we adopt a data-centric approach where we create a synthetic dataset from 197 videos, inspired by the tendencies and curiosities of students watching such videos, comprising 4982 question answer (QA) pairs. A sample data point from our dataset is presented in Figure \ref{fig:data_example}.  \\
\textbf{C2 (\textit{Study})}: To understand the qualitative requirements of the answers to questions in the dataset, we study the preferences among students to better understand the needs of the dataset and model responses. We gain novel insights about the task through this study, which we use to make changes to our synthetic dataset and evaluation methods. \\
\textbf{C3 (\textit{Evaluation metrics})}: In order to evaluate the long-form text generated with respect to the reference answers, we not only employ existing text-based metrics, but also qualitative metrics inspired by our human study (C2), which provide more nuanced insights on the performance of the models. \\
\textbf{C4 (\textit{Benchmarking)}}: We design appropriate context retrieval pipelines and benchmark our dataset across 6 state of the art MLLMs of various sizes, and analyze our data to provide useful insights and future directions of work on this task. We try to answer RQ1 and RQ2, and study the challenging nature of the task through this process.

\section{Related Work}
\subsection{Educational Video QA Datasets}

\begin{table}[!htbp]
\centering
\normalsize
\resizebox{\columnwidth}{!}{%
\setlength{\tabcolsep}{2.5pt}
\renewcommand{\arraystretch}{1.5}  
\begin{tabular}{lllll}
\toprule
\textbf{Dataset}     & \textbf{Video Type}      & \textbf{Answer Type}       & \textbf{Avg Vid Length} & \textbf{Reasoning Type}           \\ \midrule
TutorialVQA   & Tutorial    & Open Ended & 1488 secs             & Comprehension                    \\ 
How2QA        & Tutorial                 & MCQ                         & 17.45 secs             & Comprehension                    \\ 
HowToVQA      & Tutorial                 & MCQ                         & 12.1 secs              & Comprehension                   \\ 
YTCommentQA   & Tutorial                 & Open Ended                  & 524 secs               & Knowledge                        \\ 
\textbf{EduVidQA}      & \textbf{Lecture}      & \textbf{Open Ended}                  & \textbf{4054 secs}              & \textbf{Evaluation}        \\ \bottomrule
\end{tabular}%
}
\caption{Summary of datasets with their video types, answer types, average video lengths, and reasoning types. In the \textit{Reasoning Type} column, we indicate the highest level of reasoning according to the Bloom's Taxonomy \cite{krathwohl2002revision}, required to answer the questions. The average video length of EduVidQA represents that of the real-world test set.}
\label{tab:dataset-summary}
\end{table}


QA from educational videos has been explored in various settings. TutorialVQA \cite{colas2019tutorialvqa} focuses on verbose, narrative instructional videos. How2QA \cite{li2020hero} and HowToVQA \cite{yang2021just} leverage instructional videos, with the latter generating synthetic QA pairs. YTCommentQA \cite{yang2024ytcommentqa} examines questions from YouTube comments, also on instructional videos. While automated QA has been studied using traditional NLP \cite{cao2005automated, repp2008question}, no long-form QA datasets exist to benchmark educational video QA tasks. EduVidQA introduces a pedagogically designed QA dataset based on lecture videos, posing unique challenges for state-of-the-art MLLMs. Table \ref{tab:dataset-summary} and Figure \ref{fig:data_example} compare it with prior work.

\subsection{Long Text Evaluation}  

Traditional n-gram metrics like BLEU \cite{papineni2002bleu}, ROUGE \cite{lin2004rouge}, and METEOR \cite{banerjee2005meteor} have been used for summarization and translation tasks but lack human interpretability, a growing need in LLM benchmarking practices. Model-based metrics such as BERTScore \cite{zhang2019bertscore} and Entailment Score \cite{ray2024ervqa} offer semantic evaluations, while LLMs are increasingly used for reference-based assessment \cite{li2024salad, zhang2024evaluating}. Token overlap methods have been explored for correctness and faithfulness evaluation \cite{adlakha2023evaluating}, where correctness is defined as satisfying the user's information need. A similar idea, corresponding to factual precision and recall has been explored via the LLM-based FactQA metrics \cite{fernandez2024syllabusqa} which borrows its general idea from Factscore \cite{min2023factscore}. Qualitative metrics have also been explored in various works based on multi-faceted answer quality \cite{chen2023beyond}, human feedback and helpfulness \cite{ouyang2022training}, dialogue response quality\cite{thoppilan2022lamda} and comprehensive manual evaluation \cite{zhang2024llmeval}. In this work, we use a set of pedagogically relevant qualitative metrics, motivated by our preference study, which gives a nuanced evaluation of our benchmarking models. \\
Additionally, we discuss the related work that led us to explore the use of synthetic data for finetuning MLLMs in Appendix B. 
\section{EduVidQA Dataset}
The EduVidQA dataset contains 5,252 QA pairs from educational lecture videos, including 270 manually curated and verified pairs, with the rest synthetically generated from course documentation. Both real and synthetic data involve domain experts, defined as: (1) Current graduate student (PhD/Masters), (2) Having at least 1 year of research experience in the subject or a related domain, (3) Having teaching assistant experience in a related course. These criteria apply to all domain experts involved in this work.

\subsection{Creation of Real-World Dataset}

To analyze QA pairs from YouTube videos, we collected 145 long videos from 7 Computer Science and AI courses with active comment sections and available domain experts (full course list is in Appendix C.1.1). We then extracted relevant QA pairs through the following process (with examples from different stages of the process are shown and discussed in Appendix C.1.2): \\ 
\textbf{1) Regular expression-based filtering:} Using the \texttt{python-youtube} library\footnote{\url{https://pypi.org/project/python-youtube/}}, we retrieved $100$ comments per video, identifying $1440$ question-containing comments via regex. These were essentially comments which contained question marks.  \\
\textbf{2) Manual filtering of questions:} Out of the questions filtered, many were rhetorical, off-topic, or meta-discussions (check Appendix C.1.2). Accordingly, 'useful questions' were defined as those having the following properties: \\
\textit{a) Knowledge-seeking quality:} Out of the filtered questions, many were rhetorical, off-topic, or meta-discussions. We define ‘knowledge-seeking questions’ as those requiring domain expertise. Manual filtering yielded 702 such questions.  \\
\textit{b) Temporal grounding:} Many relevant questions included timestamps referring to specific video segments. We identified $270$ timestamped questions and manually added timestamps where missing. The heuristics for selecting an appropriate timestamp are discussed in Appendix C.1.3.  \\
\textbf{3) Expert verification and QA correction:}  
While some questions had expert verified answers in the comments, only 30/270 were fully answered, 76/270 had partial answers, and 164/270 had none. Extraneous text was removed, and domain experts manually answered, corrected, and verified all the QA pairs, wherever required. Experts were asked to refer to textbooks and online reference materials to aid this process. However, \textit{the use of LLMs was explicitly prohibited}.  This dataset serves as a real-world benchmark for our experiments ( hereby referred to as the ‘real-world test set’ or 'real-world set'). A detailed visualization of this dataset creation process is given in Appendix C. We also discuss visual dependence of the questions in Appendix C.3.3. We only manage to collect 270 QA pairs through this process as the annotation and manual verification is an expensive and lengthy process. Hence, we also create a synthetic dataset which is representative of the real-world samples, which we keep in the test set.

\subsection{Creation of Synthetic Dataset}

Zero-shot experiments on the real-world test set (Table \ref{tab:results}) highlighted the complexity of answering student questions, with even closed-source models like GPT-4o and Gemini 1.5 performing quite poorly. Since real-world QA collection is costly, we adopted a synthetic data-centric approach using the following steps (details in Appendix C.2):  

\noindent \textbf{1) QA pair generation using GPT-4o:} Instead of sampling video frames and using auto-generated video transcripts to generate QA pairs, we use the manually annotated transcripts available from National Programme on Technology Enhanced Learning (NPTEL)\footnote{\url{https://nptel.ac.in/}} instead (See Appendix C.2.2 for justification and more details). We choose 3 courses from reputed universities (course list in Appendix C.2.1) matching the domains of the real-world set, and initially generate 7530 QA pairs from 199 videos.  \\
\textbf{2) Filtering timestamp-less questions:} From the generated questions, we discard questions without timestamps using regex (matching the pattern observed in Section 3.1), leaving 6,546 QA pairs.  \\
\textbf{3) Adversarial Refinement:} To ensure questions require context from the video, we removed those answerable without context using GPT-4o and an entailment score threshold of 0.65 \cite{ray2024ervqa}, reducing the dataset to 4,982 QA pairs. A similar approach has been shown in \cite{rawal2024cinepile}. This process is described in detail in Appendix C.2.4. \\
\textbf{4) Timestamp accuracy:} GPT-4o assigned timestamps based on the key frames in the NPTEL transcript pdfs, which were fairly accurate. However, a manual check on 100 samples (using the process described in Appendix C.1.3) found an average absolute difference of 35.4s, which we consider while deciding upon a context window for our benchmarking experiments.  \\
\textbf{5) Bloom’s taxonomy tagging:} We then auto-tagged questions with their respective Bloom's Taxonomy \cite{forehand2010bloom} question tags (definition and process in Appendix C.2.3)  using GPT4o1-mini (using few-shot prompting) and validated 100 samples manually, obtaining a high agreement (Cohen’s Kappa \cite{hsu2003interrater} = 0.83). Since the disagreement in tags occurred between related classes within the taxonomy, we grouped them in the following way with \textit{question difficulty} tags as follows: \textbf{\textit{easy}} (Knowledge, Comprehension), \textbf{\textit{medium}} (Application, Analysis), and \textbf{\textit{hard}} (Synthesis, Evaluation).  \\
\textbf{6) Answer editing by difficulty:} Our Qualitative Process (Section 4) revealed that `application' and `analysis' level questions (`medium' difficulty) required clearer answers. Enhancements to the dataset were made accordingly using an additional preference study (See full process in Appendix D.2).  \\
Finally, we split the dataset into train (3,908 QA pairs, 78\%) and test (1,074 QA pairs, 22\%), using the latter as a synthetic test set for benchmarking.  \\
This process is also explained in detail through Figure \ref{fig:synthetic} in Appendix C.2.
\subsection{Quality Assurance and Dataset Statistics}

\begin{table}[t]
\centering
\setlength{\tabcolsep}{1.5pt}
\renewcommand{\arraystretch}{1.2}
\small 
\begin{tabular}{lcc}
\toprule
 & \textbf{Synthetic} & \textbf{Real-world} \\
\midrule
Total number of QA pairs & 4982 & 270 \\
Total number of videos & 197 & 99 \\
Avg. number of words per Q & 37.12 & 22.01 \\
Avg. number of words per A & 122.71 & 44.57 \\
Avg. number of words in transcripts & 2971.50 & 10616.11 \\
Avg. length of videos (sec) & 1302.96 & 4054.79 \\
\bottomrule
\end{tabular}
\caption{EduVidQA Dataset Statistics}
\label{tab:eduvidqa_stats}
\end{table}

\textbf{Synthetic Data Quality Assurance:} While the real-world test set is entirely expert-verified and curated, we ensure the synthetic data also meets real-world quality. Following \citet{whitehouse2023llm}, two graduate students evaluated 500 sampled QA pairs (10\% of the entire synthetic set) on \textbf{\textit{Question Naturalness (QN)}} and \textbf{\textit{Answer Soundness (AS)}}.  
QN is assessed based on: (i) Student-like curiosity, (ii) Content relevance, (iii) Logical validity. A binary label is assigned if all criteria are met. AS is evaluated on: (i) Logical soundness, (ii) Factual correctness, and a similar binary label is annotated. 
Both annotators agreed on AS for all answers. Annotator 1 found QN valid for 98\% of samples, while Annotator 2 found QN valid for 97\%, with only 1\% common disagreement on QN. The common disagreement was primarily due to the questions that were referencing the content of the lecture at a meta level (Eg: \textit{"What is the purpose of teaching this topic?"}). The authors skimmed through the data and found only 6 such samples which were removed from the synthetic set. These results indicate that the synthetic data maintains high qualitative standards. The final data statistics can be found in Table \ref{tab:dataset-summary}. We also compare the real-world set and synthetic set in terms of difficulty tag distributions and qualitative aspects in Appendix C.3, and visual dependence of questions in Appendix C.4.   
\section{Qualitative Preference Study}

Since we are concerned with providing answers to students, we find it imperative that we take the qualitative preferences of the students into account, while creating our dataset, as well as evaluating the models. We draw inspiration from previous works \cite{cross1988classroom, paul2013critical, puech2024towards} to decide the qualitative metrics to take into consideration and define them as: \\
\textbf{\textit{1) Clarity:}} Simplifies complex terms, structures explanations logically, and avoids ambiguity. \\
\textbf{\textit{2) Depth:}} Explores underlying concepts and provides sufficient detail without overwhelming the student. \\
\textbf{\textit{3) Encouraging Critical Thinking (ECT):}} Prompts further inquiry, discusses alternatives, and offers open-ended suggestions. \\
\textbf{\textit{4) Conciseness:}} Keeps the answer precise, relevant, and free of unnecessary details. \\
\textbf{\textit{5) Uses Pedagogical Techniques (UPT):}} Uses examples and step-by-step explanations to enhance understanding. \\
Depth and Conciseness were purposely chosen as different qualities, although both depend on the same observable features (text length, detailing, etc) in order to clearly understand the preferences among the students. We conduct a preliminary study to understand the qualitative preferences of students in our dataset’s answers.

\subsection{Study Details}
\begin{figure}[!t]
    \centering
    \includegraphics[width=0.7\columnwidth]{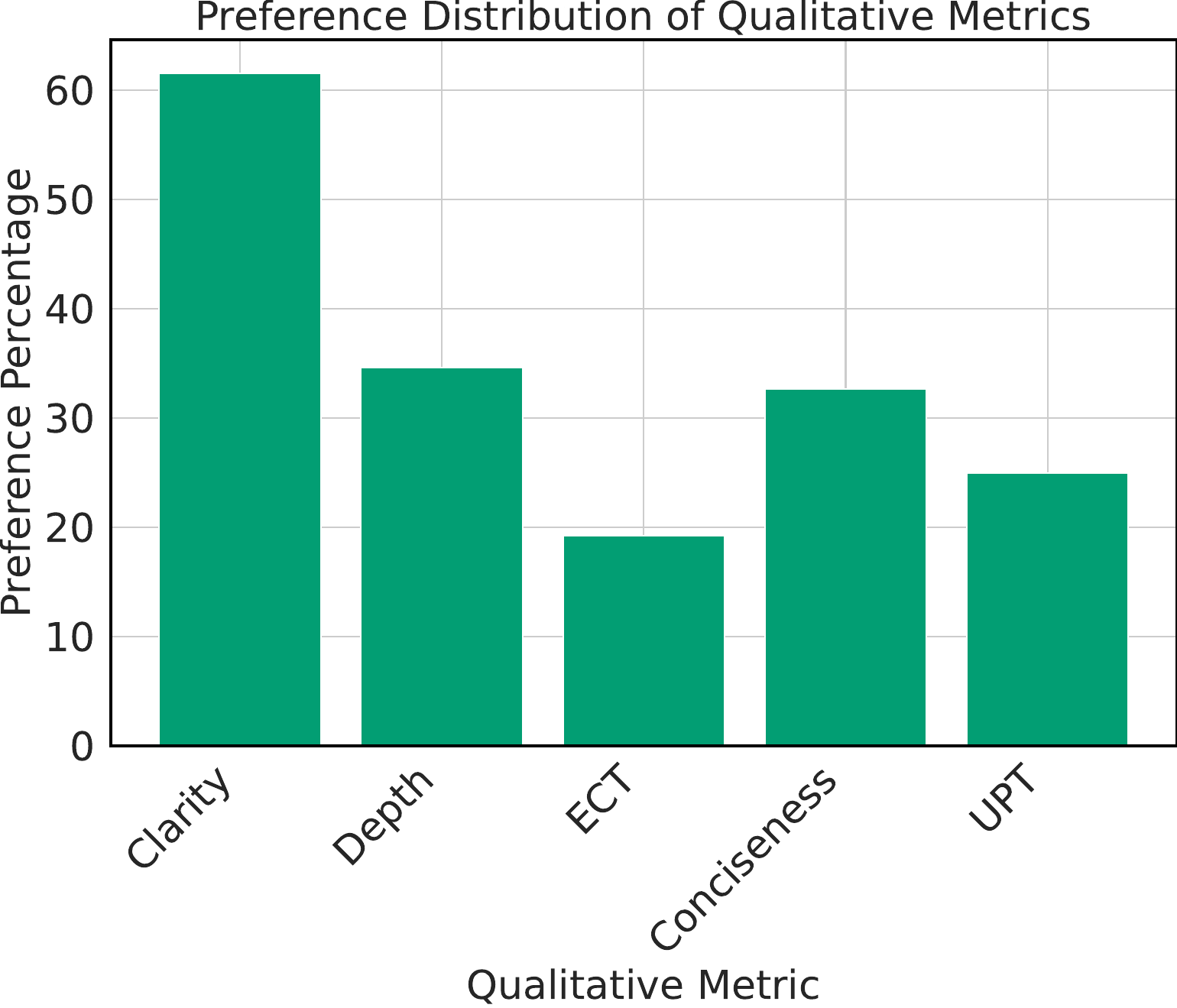}
    \caption{Results of Study 1 on the Importance of Different Quality Metrics.}
    \label{fig:qstudy1}
\end{figure} 
\textbf{Setting:} We provided 10 university students (7 undergraduates, 3 graduates) with 5 random questions from the synthetic dataset. Each question had two answers—one from the dataset and one explicitly modified. The modifications were done manually, where answers were appended with sentences to enhance UPT and ECT, shortened, or made more verbose (see example data point in Appendix D.1). Students selected their preferred answer and listed the qualities influencing their choice, from the list of qualities under consideration. \\ 
\textbf{Results:} Figure \ref{fig:qstudy1} shows that Clarity was preferred in over 60\% of responses (10 students × 5 questions). ECT and UPT were chosen less frequently. Students preferred alternative answers for ‘medium’ difficulty questions, prompting a second study (See Table \ref{tab:study_1_example} in Appendix D.2). Depth and Conciseness were selected in ~40\% of responses, with undergraduates favoring Depth and graduates preferring Conciseness (more details in Appendix D.1). As student expertise is not the focus here, we do not consider these in the subsequent work. 

\subsection{Implications and Considerations}
Leaving out Depth and Conciseness due to their dependence on the level of expertise of the student, we consider the remaining 3 qualities in the rest of this work, owing to them getting considerable preferences among the students. We identify Clarity as a metric widely preferred, being of utmost importance, needing to be maximized in the generated answers. ECT and UPT are slightly more complex. These qualities are more dependent on the type of question, and the answer it warrants. We use this information to design our evaluation metrics appropriately in Section 5.2.

\section{Benchmarking}
To show the complexity of the task, we evaluate our real-world test set and synthetic test set on various open source and closed MLLMs, and evaluate their performance. We take both Video Large Language Models (Video LLMs) as well as Large Vision Language Models (LVLMs) into consideration for our experiments. 
\subsection{Baselines}
\textbf{Video LLMs: } We use 2 state of the art open-source Video LLMs to benchmark our dataset. The mPLUG-Owl3 8B model \cite{ye2024mplug} uses hyper-attention blocks to semantically integrate the frames from the video and the text query. It uses the Siglip-400m visual encoder \cite{zhai2023sigmoid} and Qwen2 \cite{yang2024qwen2} as the language model. We also use the Video LlaVA-7B model \cite{lin2023video}, which uses LanguageBind encoders \cite{zhu2023languagebind} and a Vicuna decoder. \\
\textbf{LVLMs: } For benchmarking using LVLMs, we use both open-source and closed models. The closed models comprise of Gemini 1.5 \cite{team2024gemini} and GPT4o \cite{hurst2024gpt}. We also use open source LVLMs such as Qwen2VL-7B \cite{wang2024qwen2}, which has a  675M parameter Vision Transformer (ViT) \cite{dosovitskiy2020image} as the image encoder, and Qwen2 as the LLM decoder. We also experiment on the Llava 1.5-13B model \cite{liu2024visual}, having a ViT large based visual encoder, and a Vicuna 13B LLM decoder \cite{chiang2023vicuna}. The prompts are given in Appendix E.1.

\subsection{Evaluation Metrics}
We evaluate our models with both text-based and qualitative metrics that help us to get a nuanced perspective on the performance of the models.
\subsubsection{Text-based Evaluation Metrics: } 
For free-form text generation, the NLP community has predominantly used n-gram based metrics. We show results using BLEU-1 \cite{papineni2002bleu}, ROUGE-L \cite{lin2004rouge} and METEOR \cite{banerjee2005meteor}. We also evaluate using a semantic metric, Entailment Score (ES) \cite{ray2024ervqa}, which checks the probability of the generated text being entailed within the reference text. We also use LLM-based FactQA Precision and Recall metrics \cite{fernandez2024syllabusqa}, that give a clearer picture of the notions of correctness and completeness of the generated answers with respect to the ground truth.  However, these scores  do not provide much information about the quality of the generated text. Hence, we design a set of new metrics based on the Qualitative Preference Study, that provide a more intuitive and interpretable evaluation of the free-form generated text. 
\subsubsection{Qualitative Metrics }
\begin{table*}[!htbp]
\centering
\small
\renewcommand{\arraystretch}{1.5}
\setlength{\tabcolsep}{0.5em}
\begin{tabular}{p{0.05\textwidth}p{0.30\textwidth}p{0.30\textwidth}p{0.30\textwidth}}
\toprule
\textbf{Score} & \textbf{Clarity Scale} & \textbf{ECT Scale} & \textbf{UPT Scale} \\
\midrule
1 & >=2 jargon terms without explanation, and >=2 incoherent transitions. & No questions, no alternatives, purely factual. & Pure explanation without any example or breakdown.\\

2 & >= 1 jargon term unexplained and at least 1 logical jump or ambiguous phrasing. & Includes 1 suggestive or reflective phrase, but no actual open-ended question. & 1 brief example or partial list of steps, lacking clarity. \\

3 & Mostly clear, but 1–2 minor issues: one ambiguous phrase or slightly choppy flow. & Contains 1 open-ended question or 1 alternative method/viewpoint. & 1 complete example or full step list present, but not both. \\

4 & All terms explained, clear flow, no ambiguity except possibly 1 unclear phrase. & >=2 open-ended prompts or multiple viewpoints briefly compared. & >=2 teaching techniques used (e.g., example + step list), with moderate clarity. \\

5 & No unexplained jargon, consistent logical flow, zero ambiguity. & >=2 open-ended questions + explicit invitation to explore further. & >=3 techniques (e.g., example, analogy, visual mention), all clear and complete. \\
\bottomrule
\end{tabular}
\caption{Likert Scales for Clarity, ECT and UPT, used for both human annotations, and for prompting GPT-4o, in order to evaluate the models qualitatively.}
\label{tab:likert_scale_1}
\end{table*}

\textbf{a) Motivation:} Our Qualitative Preference Study shows that students give a lot of importance on the qualitative metrics explored through the study - with Clarity being the most important aspect, followed by UPT and then ECT. However, getting these qualities manually annotated for all generations can be an expensive and a time-consuming process. Hence, we introduce LLM (GPT-4o) based methodologies to get a quantitative estimate of these qualities with the help of Likert Scales \cite{joshi2015likert}. Before we did this, we wanted to ground our interpretations to that of actual students, and validate our Likert Scale. \\
\textbf{b) Human Annotation Study:} \\
To ensure validity of our LLM-based metrics, we have 2 student annotators (graduates) to annotate the quality scores according to the Likert Scale defined in Table \ref{tab:likert_scale_1}, for 320 data points from the test set (50\% split of real-world and synthetic). We also experiment with another version of this scale, with a more relaxed objective definitions (shown in Appendix E.2.3), but we settle on this scale due to this being easier to annotate and having a highly positive inter-annotator Spearman's $\rho$  \cite{de2016comparing} of 0.7029 for Clarity, 0.7227 for ECT and  0.7815 (p values were less than 1e-7 for all, ensuring high statistical significance). We justify the usage of Spearman's $\rho$ from previous works exploring this idea \cite{sullivan2013analyzing}. We also observe a relatively high Cohen's Kappa score of 0.6966, 0.6715 and 0.7564 for Clarity, ECT and UPT respectively. Annotation instructions and term definitions have been discussed in detail in Appendix E.2.2. \\
\textbf{c) LLM Prompting and Human Agreement: } \\
For getting the qualitative scores from the answers in a reference-free way, we add the definition of the objective, the Likert Scale output format, in-context examples for each score on the scale and finally a query prompt. The full prompts are shown in Appendix E.2.2. We then use GPT-4o to obtain the scores. For the scores for which both human annotators agreed (207/320 for Clarity, 142/320 for ECT and 196/320 for UPT), we get scores from GPT and find the Cohen's Kappa to be 0.7162 for Clarity, 0.4586 for ECT and 0.8486 for UPT, showing moderate to high agreement between GPT-4o scores and human annotations. The reason the correlation is moderate for ECT, is possibly because of a higher chance of misunderstanding of nuanced key terms such as \textit{suggestive/reflective phrase}, in spite of refined definitions provided to the annotators (described in Appendix E). We also get impressive Spearman's $\rho$ scores of 0.4071 for Clarity, 0.8709 for ECT and 0.8689 for UPT, thus justifying our usage of this Likert Scale and prompting method. \\
\textbf{d) Evaluation Methodology: } \\
From the Qualitative Preference Study, we can infer Clarity to be a quality which is desirable in all answers, whereas ECT and UPT are more dependent on the question and its complexity. Hence, a good model is one that produces answers with a high average Clarity score, and one that has ECT and UPT scores closest to the ground truth score, assuming the answer content is factually correct. Hence, for ECT and UPT, we use Spearman's $\rho$ with respect to the ground truth scores to determine the performance of the models.

\subsection{Experimental Settings}
\textbf{Task setting: } To perform our experiments, we sample the reference frame from the timestamp in case of LVLMs. In case of Video LLMs, we sample the reference frame along with 14 frames before and 15 frames after the reference (30 frames in total) from a context window of 120 secs before and after the reference time stamp. Hence, the total context size around the reference time stamp is 4 mins. Similarly, we also feed the text transcript or audio (for Video Llama 2) from the 4 min context window to the models, extracted from the \texttt{python-youtube} API. This was done according to the manual check done in Section 3.2 which found an average absolute difference of 35.4s between the dataset timestamps and the manual timestamps. We consider a wider range, to cover for the worst-case scenario where he subject matter of the question is beyond the average absolute difference by a considerable difference. Please note that the timestamp from the question is parsed to get the frames and transcripts from the video. \\
\textbf{Experimental Setup: }All models were run on 2 NVIDIA L40 GPUs for 10 epochs (early stopping was used to account for early convergence). We quantize all models to \texttt{bfloat16}, and conduct LoRA finetuning with \texttt{r = 64}. We tune the learning rate to 2e-5 for Llava and 1e-5 for all other models. We also fix the maximum token length to 256 for all models. 
\section{Results and Analysis}
\begin{table*}[!ht]
\resizebox{\textwidth}{!}{
\begin{tabular}{l|cccccc|cccccc}
\toprule
 & \multicolumn{6}{c}{\textbf{Synthetic Test Set}} & \multicolumn{6}{c}{\textbf{Real World Test Set}} \\
\midrule
\textbf{Model} & \textbf{BLEU} & \textbf{ROUGE-L} & \textbf{METEOR} & \textbf{Entail} & \textbf{FQA-P} & \textbf{FQA-R} & \textbf{BLEU} & \textbf{ROUGE-L} & \textbf{METEOR} & \textbf{Entail} & \textbf{FQA-P} & \textbf{FQA-R} \\
\midrule
\textbf{Finetuned Video LLMs} & & & & & & & & & & & & \\
Video LlaVA 7B &25.50 &29.10 &21.88 &11.40 &52.21 &42.53 &12.47 &21.84 &17.55 &13.09 &45.32 &21.60 \\
mPLUG Owl 3 8B* & 33.58 & 32.17 & 23.74 & 20.08 & 62.68 & 50.87 & 18.17 & 31.79 & 19.64 & 17.84 & 57.33 & \textbf{29.34} \\
\midrule
\textbf{Finetuned VLMs} & & & & & & & & & & & & \\
Qwen VL 7B & 25.84 & 28.87 & 27.21 & 18.40 & 67.21 & 64.81 & \textbf{18.36} & \textbf{33.00} & 19.22 & 18.80 & 60.83 & 25.48 \\
Llava 13B* &32.19 &22.86 &32.22 &20.68 &61.35 &51.03 &17.16 &20.03 &\textbf{31.41} &18.06 &61.35 &27.29 \\
\midrule
\textbf{Closed VLMs} & & & & & & & & & & & & \\
GPT4o* &- &- &- &- &- &- & 15.03 & 32.87 & 22.46 & \textbf{29.26} & \textbf{64.49} & 25.98 \\
Gemini 1.5 &31.07 &28.62 &32.55 &29.77 &73.29 &50.98 & 13.98 & 30.76 & 21.25 & 21.13 & 63.27 & 25.15 \\
\bottomrule
\end{tabular}}
\caption{Benchmarking Results on the synthetic test set and real-world test set using text-based metrics. \textbf{Bold} figures indicate best performance across all models for the real-world test set. * indicates best performing model across each model category, according to FactQA-Precision (FQA-P) and FactQA-Recall (FQA-R) on the real-world test set.}
\label{tab:results}
\end{table*}

\begin{table*}[!ht]
\centering
\resizebox{\textwidth}{!}{
\begin{tabular}{l|cccccc|cccccc}
\toprule
 & \multicolumn{6}{c}{\textbf{Synthetic Test Set}} & \multicolumn{6}{c}{\textbf{Real World Test Set}} \\
\midrule
\textbf{Model} & \textbf{Clarity Avg} & \textbf{Clarity AD} & \textbf{ECT Avg} & \textbf{ECT $\rho$} & \textbf{UPT Avg} & \textbf{UPT $\rho$} & \textbf{Clarity Avg} & \textbf{Clarity AD} & \textbf{ECT Avg} & \textbf{ECT $\rho$} & \textbf{UPT Avg} & \textbf{UPT $\rho$} \\
\midrule
\textbf{Closed VLMs} & & & & & & & & & & & & \\
GPT4o & - & - & - & - & - & - & \textbf{4.1585} & 0.5727 & 1.8869 & 0.1378 & 1.3491 & 0.1987 \\
\midrule
\textbf{VLMs} & & & & & & & & & & & & \\
Llava 13B (0 shot) & 3.4247 & 1.2818 & 1.2310 & 0.3328 & 1.6477 & 0.2699 & 2.8834 & 0.7023 & 1.0840 & 0.1223 & 1.3435 & 0.1037 \\
Llava 13B (SFT) & 3.8484 & 0.8580 & 1.2432 & 0.3668 & 1.6949 & 0.3265 & 3.1901 & 0.3956 & 1.1472 & \textbf{0.2413} & 1.2944 & \textbf{0.2459} \\
\midrule
\textbf{Video LLMs} & & & & & & & & & & & & \\
mPLUG Owl 3 8B (0 shot) & 3.4672 & 1.2393 & 1.0737 & 0.2750 & 1.1844 & 0.2889 & 3.2682 & 0.3175 & 1.0548 & 0.0411 & 1.0670 & -0.0083 \\
mPLUG Owl 3 8B (SFT) & \textbf{4.3629} & \textbf{0.3436} & 1.1930 & \textbf{0.3841} & 1.4054 & \textbf{0.3376} & 3.8841 & \textbf{0.2983} & 1.3780 & 0.2388 & 1.2378 & 0.1958 \\
\bottomrule
\end{tabular}}
\caption{Qualitative benchmarking results on \textit{best performing models}. \textit{AD} denotes the absolute difference between the average scores of the ground truth data and the generated data, \textit{Avg} denotes average across all test samples and $\rho$ denotes the Spearman's Correlation Coefficient. Average Ground-truth scores: Clarity (Real-world Avg = 3.5857; Synthetic Avg = 4.7065), ECT (Real-world Avg = 1.1775; Synthetic Avg = 1.3243), and UPT (Real-world Avg = 1.1893; Synthetic Avg = 1.5212). \textbf{Bold} figures indicate the best performing model for each qualitative metric for the synthetic and real-world sets. All Spearman's $\rho$ scores are statistically significant (p-value < 0.001).}
\label{tab:qualityresults}
\end{table*}

\begin{figure}[!t]
    \centering
    \includegraphics[width=\columnwidth]{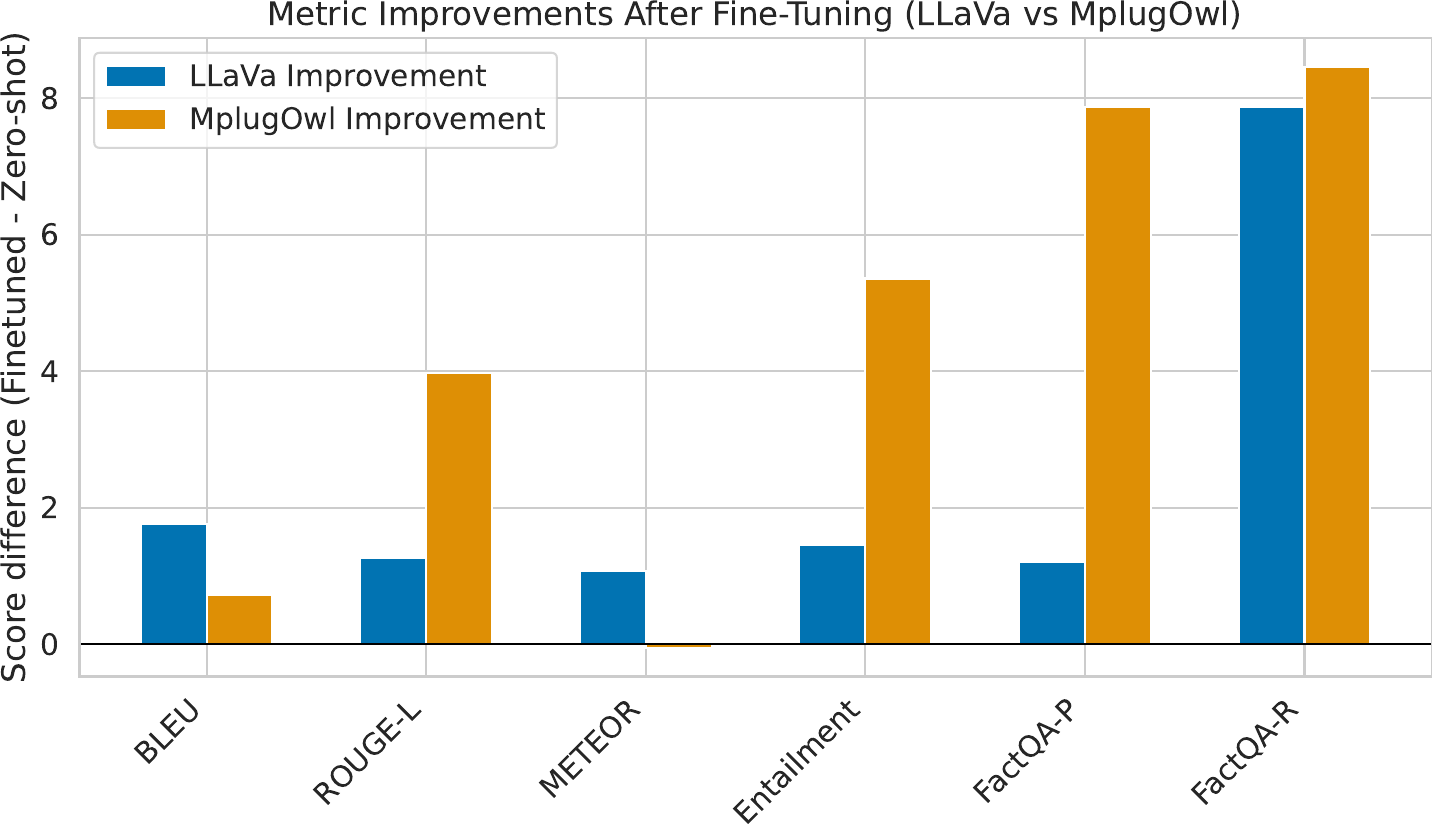}
    \caption{Improvement in scores (Finetuned metric - Zero-shot metric) across all the metrics, averaged across both test sets, for our \textit{best performing models} }
    \label{fig:improvement}    
\end{figure}

The text-based results after finetuning the baseline models are reported in Table \ref{tab:results},  and the qualitative metrics of the best models are shown in Table \ref{tab:qualityresults}. \\ 
\textbf{a) Text-based analysis:}
For the synthetic test set, our experiments show that finetuning a comparatively smaller open source model in Qwen-VL, finetuned on data from the same distribution can generate answers with higher FactQA-Recall compared to a much larger model in Gemini. Since the training data and test data are from the same distribution, we observe impressive results for the finetuned models, for the synthetic test set. In the real-world test set, we see interesting trends on finetuning smaller open source models using synthetic data. All open source models show comparable performance to closed models on most metrics, even in in terms of more interpretable metrics FactQA Precision and Recall. For the real-world test set, we see comparable metrics across all models, but for the sake of comparing, in the case of FactQA Precision and Recall, Llava-13B outperforms Qwen2-VL 7B possibly due to its difference in the number of parameters. We also see a similar trend across the Video LLMs, where mPLUG Owl 3-8B outperforms Video LlavA-7B. \textit{Hence, we consider Llava-13B and mPLUG-Owl 8B as our best performing models for other analyses. GPT-4o is the best performing closed-source model.} \\
\textbf{b) Qualitative Analysis: }
Table \ref{tab:qualityresults} shows the results for qualitative analysis on the best performing models. We use the insights from Section 5.2.2 d) to evaluate our models. It is interesting to observe that finetuned mPLUG Owl 3 8B generates answers with the highest Clarity, and has the lowest absolute difference with respect to the synthetic test set. The finetuned mPLUG Owl 3 8B also has higher ECT and UPT $\rho$ scores for the synthetic test set. For the real-world test set, although GPT4o has the highest average Clarity, finetuned mPLUG Owl 3 8B has the least absolute difference. Similarly, we observe finetuned Llava-13B to have the highest Spearman's $\rho$ with respect to the real-world dataset for ECT and UPT. \\
\textbf{c) Effects of Finetuning: }
Supervised finetuning on the synthetic data unanimously improves the model performance. After finetuning, both open-source models show improvement across all text-based metrics, as shown in Figure \ref{fig:improvement}. For the qualitative metrics, Table \ref{tab:qualityresults} also shows an improvement of metrics on supervised finetuning. Although even the best results show room for improvement, the supervised finetuning experiments show the efficacy of the synthetic training set for the task of finetuning smaller models in the range of 7B to 13B parameters to match or even better their performance compared to much larger closed models. 

\section{Conclusion and Future Work}
In conclusion, let us once again revisit the RQs we established in the Introduction, and examine our progress. Through our extensive studies and surveys, we have established a qualitative preference among students that can be generalized to a certain extent. 
Moreover, the traditional evaluation metrics used in this line of work failed to cover the nuances needed to estimate human standards of what constitutes an \textit{adequate answer}. Our qualitative metrics, based on our human study, provide a more aligned representation of these standards, which is an important step towards estimating the performance of MLLMs - a crucial resource for our research objective. Through finetuning open-source models on our synthetic data, we show that it is possible for smaller models to provide comparable, and sometimes even better responses compared to larger closed models, both in terms of quantitative and qualitative metrics. However, even the best performing models have limitations which allows room for improvement, and opens a new research direction specific to this utility. Our dataset acts as a pivotal resource for progress in this line of work. Additionally, some preliminary results in Appendix F provide reasonable directions on future progress in this task - to tailor models to reason on the basis of question objective and difficulty. Hence, through this work, we not only move closer to providing consistent and on-demand academic support for diverse student communities, but also contribute to a more inclusive learning landscape for the future.
\section*{Limitations}

\textbf{Data Annotation and Human Studies:} Our human level evaluation studies and surveys are restricted to certain demographics, due to a constraint in the availability of subjects at the time of conducting this research. We are currently exploring the possibilities of extending these studies to a wider community, to capture a more diverse range of cognitive models, and validate our results even further. Another limitation to our work is the size and coverage of the real-world dataset, also due to availability issues with specific domain experts, and due to the data annotation process being an extremely time-consuming and expensive activity. Extending this work beyond the domain of Computer Science remains our utmost priority. Although our annotators were not sourced from a particular platform, we compensated them on an hourly basis. Human study subjects were volunteers. All key student contributors have been recognized in the Acknowledgments Section for their crucial role in this work. \\
\textbf{Evaluation metrics:} Our qualitative evaluation metrics are highly dependent on the reasoning capabilities of LLMs. Although LLMs show impressive capabilities in rule-based instruction-following tasks , using an open source alternative to GPT4o might hamper the performance and accuracy of the metric. \\
\textbf{Experiment Design: }Our open source model selection for benchmarking was restricted to smaller models due to compute resource constraints. 

\section*{Ethics Statement}
1) The EduVidQA dataset is based on computer science lectures, which might not represent a diverse student population in terms of language proficiency, gender, socio-economic background, or accessibility needs. Since the annotation of the QA pairs in the real world test set is conducted by a restricted set of domain experts, it may be possible that the data reflects personal standards over broader perspectives. \\
2) The synthetic data generation process could introduce biases from pre-trained language models that might over-represent certain perspectives and exclude others. Also, if finetuned MLLMs inherit biases from pre-existing datasets, they might generate responses that are skewed towards dominant cultures, gender stereotypes, or regional academic preferences.\\
3) In the case of NPTEL content which are used for our large synthetic data creation, the videos are distributed under the Creative Commons Attribution-ShareAlike license which allows reuse of videos along with appropriate credit, which we have provided by specifying the actual video links in the dataset, and do not redistribute the data in any way. Also in the case of other videos used in our work, we follow the same approach, and align with the standard data practices followed in widely used large scale datasets like TutorialVQA \cite{colas2019tutorialvqa}. \\
4) While AI can enhance learning, over-reliance on automated answers might reduce critical thinking skills in students or replace human interaction in learning. Works like these are meant to complement, and not replace human educators who have dedicated their life to teaching students. \\
5) We acknowledge the usage of large enterprise LLMs like ChatGPT and Le Chat, as occasional writing and coding assistants when required. However, all novel contributions in this work, including metric definitions, taxonomies, studies/surveys, experiment design and evaluation metric design have been conceived independent of these LLMs, based on intuitions, experimentation and empirical observations. 

\section*{Acknowledgments}
This research was partially supported by the IIT Kharagpur Technology Innovation Hub on AI for Interdisciplinary Cyber-Physical Systems (AI4ICPS) at IIT Kharagpur (through the project titled \textit{“Using Large Language Models to Enhance Learning Efficiency and Student Engagement in Indian Education System”}). The authors would also like to acknowledge the undergraduate and graduate students from IIT Kharagpur who volunteered to take part in the human studies and surveys that helped us gain invaluable insights. Special mention must also be made for the contributions of graduate students Anubhav Dhar, Abhilash Nandy, Anindita Mandal and Argha Sen for their crucial role in creating the dataset. The authors would also like to acknowledge Ishani Mondal, a PhD student at University of Maryland, College Park, for her valuable suggestions during the initial conceptualization of the task.

\bibliography{custom}

\appendix

\section{Motivational Study Details}

\subsection{Related Studies}

Our motivational study for this work is inspired by several related works in this area corresponding to the explorations of the real world problem under consideration. While there are various studies which show the adoption of online lecture videos instead of books among Generation Z \cite{Langreo_2022, burhanli2021university}, the lack of interaction with instructors have been covered as a major drawback of video-based learning \cite{stecula2022advantages}. A preliminary form of our study also shows that only 3-7\% of learners ever participate in discussion forums \cite{he2018participating}, underscoring that more learners do not engage in visible questioning and answering. This study also find that some students are discouraged by the forum/comment environment itself – prior analyses observed that many MOOCs suffer from information overload and spam in forums, making it hard to find relevant information - a finding that is unanimously reinforced by our study. These works prompted us to explore this problem further and gain better insights into the problem. 

\subsection{Design Considerations}

\textbf{Aim: } Inspired by the studies mentioned above, we wanted to explore and provide empirical support by quantifying 1) The demand for interaction in lecture videos and 2) The deficiencies of current platforms like YouTube in enabling this interaction. We do this by designing the study to a) Collect both behavioral and attitudinal data on video learning b) Identify actionable gaps in current platforms c) Justify the need for expert-aligned QA systems d) Inform synthetic data generation and human preference evaluation.  \\
\textbf{Population Demographics: } We take insights from 52 students, of which 75.5\% are undergraduate students and 24.5\% are graduate students, through which we aim to capture a broad range of learning experiences relevant to MOOC-style lecture videos. \\
\textbf{Question Design:} We carefully design the following list of questions for our study: \\
a) How often do you use Youtube MOOC videos or other educational videos to learn new material? \\
b) Have you ever felt like asking a question about the content to the educator while watching such videos? \\
c) Do you end up asking content related questions in the Youtube comment section? For example, a clarification question based on what is being taught. \\
d) If you don’t tend to ask questions, what is the reason? \\
e) If you receive an answer, how satisfied are you with the response usually? \\
The final list of questions were finalized after an initial pilot study on 3 graduate students with research experience in Human Computer Interaction, who were also asked to provide meta-level feedback on the study. This practice is also followed in the subsequent study.

\subsection{Additional Insights from the Study}
\begin{figure*}
    \centering
    \includegraphics[width=0.95\textwidth]{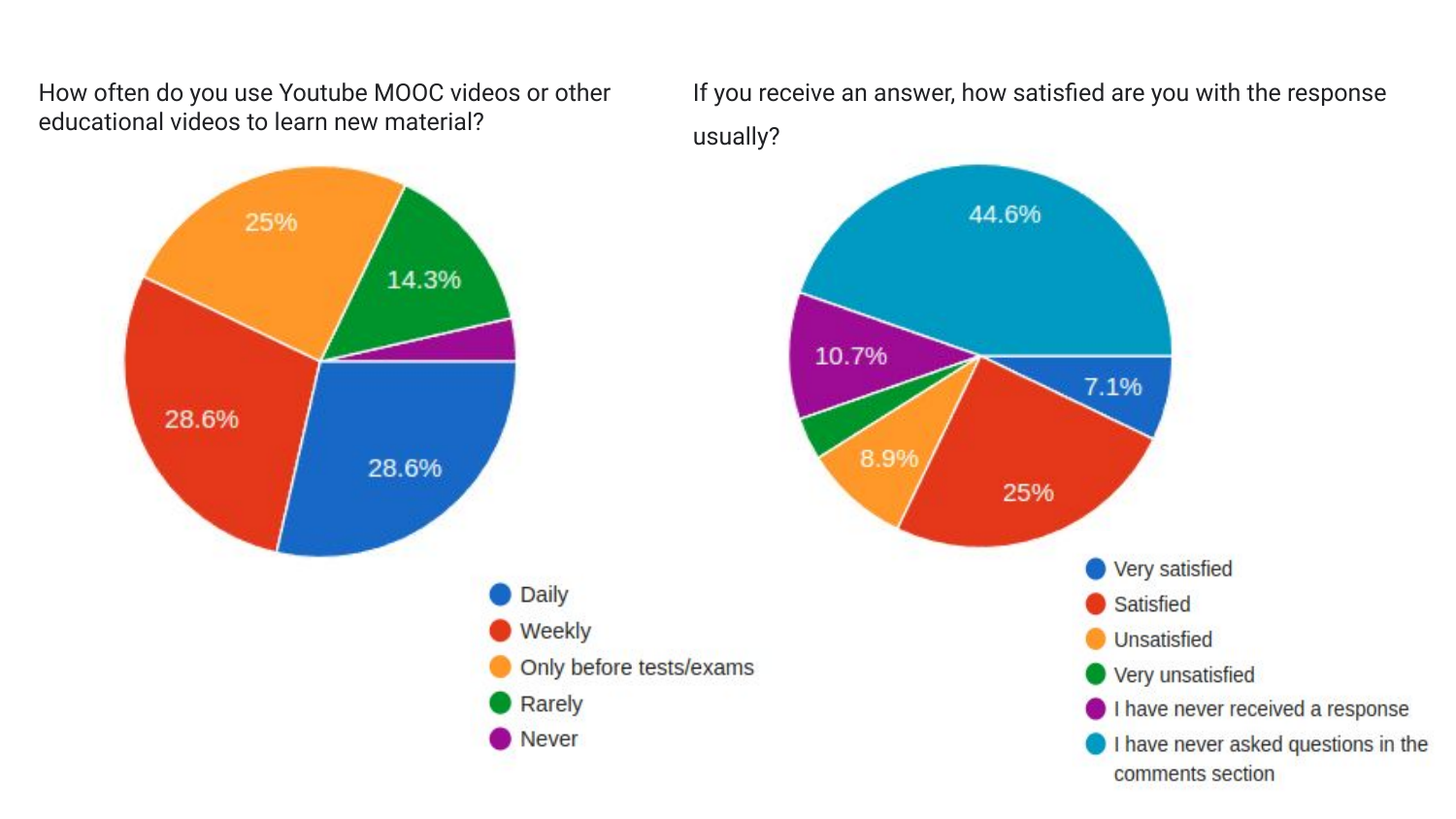}
    \caption{Some granular level data from the Motivation Study from the Introduction}
    \label{fig:motivation_extra}
\end{figure*}

As is evident from the questions, the study to draw the motivation for working on this problem, described in the paper, was conducted in more detail than what is mentioned, in order to understand student interaction patterns better and to understand the satisfaction levels of students who do ask questions. We find that a large number students watch online lecture videos daily (28.6\%), as well as weekly (28.6\%) while a large number of students (25\%) watch these videos before exams/tests. This elaborates on the relevance of the problem further, and the response to their satisfaction after receiving an answer, which shows 8.7\% of the students are unsatisfied, 10.7\% of the students are very unsatisfied, along with the large percentage of students (44.6\%) who do not usually ask questions, show the urgent need for a solution to be developed for this problem. These granular level results are shown in Figure \ref{fig:motivation_extra}.

\section{Justification for Finetuning on Synthetic Data}

The major approach explored in this paper, as a potential solution for the problem at hand, is the usage of synthetic data for finetuning the open source MLLMs under consideration. 
Synthetic data has become a cornerstone in training large language and vision-language models due to its scalability, cost-effectiveness, and potential for enhancing model capabilities. Foundational works in NLP and multimodal learning demonstrated that models trained on automatically generated QA pairs or instructions can rival those trained on human-annotated data \cite{yang2021just}. Instruction-tuning frameworks like FLAN \cite{longpre2023flan} and Self-Instruct \cite{wang2022self} have shown that LLMs can be aligned to follow diverse instructions using entirely synthetic prompt-response pairs. Recent advances, such as Orca \cite{mukherjee2023orca}, further highlight the benefit of incorporating teacher-model-generated rationales and step-by-step explanations. In the VLM space, works like LLaVA \cite{liu2024visual} uses synthetic data to teach image or video understanding at scale. While challenges like hallucination and quality control remain \cite{liu2024survey}, studies show that, with proper filtering and reasoning-focused design, synthetic datasets can significantly improve model performance in educational, instructional, and long-form QA tasks. We believe our highly tailored approach to generating synthetic QA pairs provide a larger representative sample space for the task, as well as provide valuable future insights into the usage of synthetic data for pedagogical applications. 
\section{Dataset Creation Details}

\subsection{Real-world Data Creation Details}
\begin{figure*}
    \centering
    \includegraphics[width=0.95\textwidth]{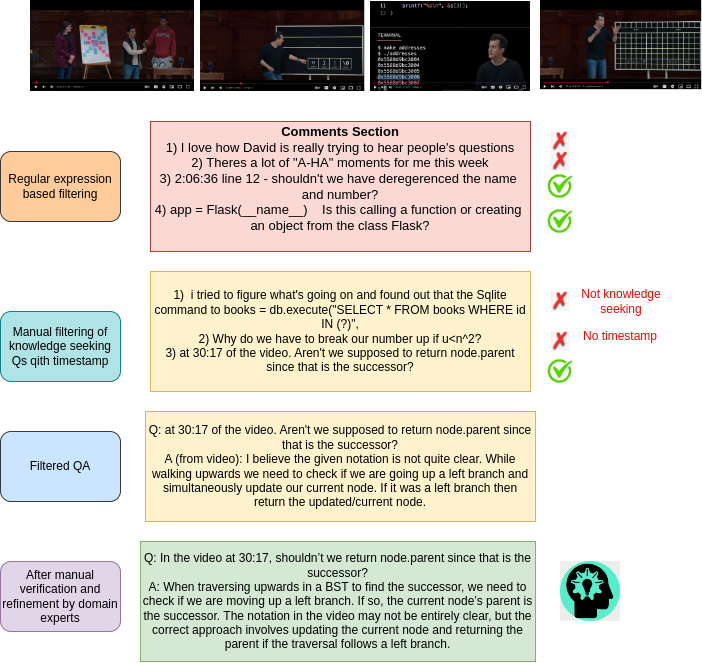}
    \caption{Real world dataset creation illustrated through examples}
    \label{fig:real_world}
\end{figure*}
The real-world dataset has been created from Youtube videos of online courses from various well-known Computer Science courses. The comments from these videos were taken and filtered using careful heuristics mentioned in Section 3 of the paper. The diagram of the entire annotation pipeline is presented in Figure \ref{fig:real_world}.

\subsubsection{List of courses}
The list of courses under consideration are as follows (Course Name/ Course Code/ Number of Videos / University Name): \\
1) Machine Learning/ CS229/ 20/ Stanford University \\
2) Introduction to Computer Science/ CS50/ 13/ Harvard University \\
3) Introduction to Algorithms/ 6.006/ 32/ Massachusetts Institute of Technology \\
4) Natural Language Processing/ CS231N/ 15/ Stanford University \\
5) Deep Learning/ 11-785/ 24/ Carnegie Mellon University \\
6) Deep Learning/ 6.S191/ 8/ Massachusetts Institute of Technology \\
7) Deep Learning/ DS-GA 1008/ 33/ New York University

\subsubsection{Examples of Filtering and Verification}
\begin{table*}[!htbp]
\centering
\small
\renewcommand{\arraystretch}{1.5}
\begin{tabular}{|l|p{0.4\textwidth}|p{0.30\textwidth}|}
\hline
\textbf{Filtering Type} & \textbf{Comment / QA Pair} & \textbf{Annotation} \\
\hline
Regular Expression Filtering & “Can you explain why we need ReLU after every convolution?” & Accept (contains “?”) \\
Regular Expression Filtering & “Awesome explanation of gradient descent!” & Reject (no “?”) \\
\hline
Manual Filtering & “Why does every course use MNIST examples? Isn’t it outdated now?” & Rhetorical \\
Manual Filtering & “This is pretty cool, right?” & Off-topic \\
Manual Filtering & “Sir, please upload the next lecture soon?” & Meta discussion \\
Manual Filtering & “At 12:45, is the backpropagation update wrong? I thought the derivative of tanh is different.” & Accept \\
\hline
Expert Verification & 
\textbf{Q:} “What is cross-validation in CNN?” \newline
\textbf{Corrected:} “What is cross-validation, and how can it be used while training CNNs?” 
& Noisy Question Corrected \\
Expert Verification & 
\textbf{Q:} “How is the cost function minimized using gradient descent?” \newline
\textbf{A:} “You add the gradient to the weights to reduce the loss.” \newline
\textbf{Corrected A:} “You subtract the gradient multiplied by the learning rate from the weights, which gradually reduces the cost function.” 
& Noisy Answer Corrected \\
\hline
\end{tabular}
\caption{Examples of filtering techniques and expert corrections applied to student comments/questions during real-world dataset annotation.}
\label{tab:real_world_examples_table}
\end{table*}

We show examples of filtering and expert verification for the real-world dataset in Table \ref{tab:real_world_examples_table}. These are explicitly shown to make the reader understand the various types of annotation in the filtering process, and the design choices made for the data. The different types of filtering shown are regular expression filtering and manual filtering.

\subsubsection{Manually adding Timestamps}
The process of manually adding timestamps to the QA pairs which did not have any associated timestamps followed the following heuristics: \\
1) The experts were asked to first go the video link and search for keywords from the question in the official transcript available on the site. \\
2) Youtube provides a list of timestamps with the exact keyword match. \\
3) The expert was then instructed to watch the most suitable timestamps according to the candidate timestamp list, and choose the timestamp where the content most matched the subject matter of the question. \\
These heuristics were decided upon experentialy, after small pilot annotations conducted by the authors.

\subsection{Synthetic Dataset Creation Details}
\begin{figure*}
    \centering
    \includegraphics[width=0.95\textwidth]{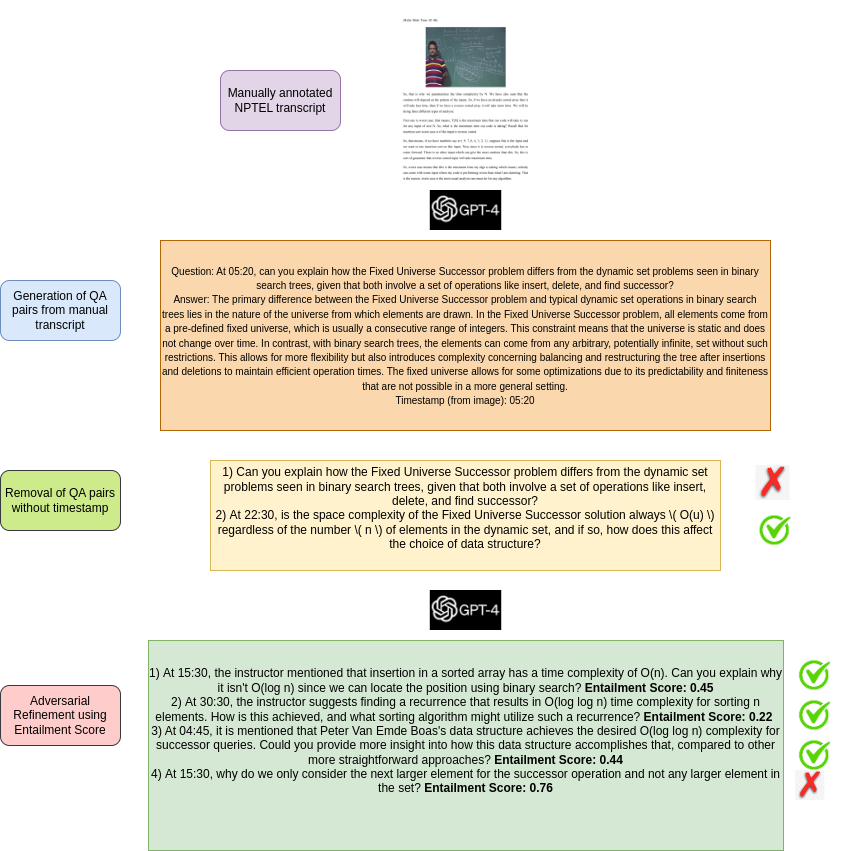}
    \caption{Synthetic dataset creation illustrated through examples}
    \label{fig:synthetic}
\end{figure*}
The entire process of synthetic data creation is visualized in Figure \ref{fig:synthetic}. We also provide the following additional details of the process, which aim to give more clarity behind our data creation pipeline.

\subsubsection{Course List}
We choose 3 NPTEL courses to generate synthetic questions from, keeping in mind diversity in technical difficulty and type of content, and availability of transcripts. The following courses were chosen as a result (Course Name/ Course Code/ Number of Videos / University Name): \\
1) Introduction to Algorithms and Analysis/ noc20-cs10/ 54/ Indian Institute of Technology, Kharagpur \\
2) Deep Learning - Part 1/ CS7015/ 118/ Indian Institute of Technology, Madras \\
3) Computer Networks and Internet Protocol/ noc22-cs19/ 52/ Indian Institute of Technology, Kharagpur \\
Of these courses, 199 videos were deemed suitable with adequate educational content and available pdf transcripts.

\subsubsection{Advantage of using NPTEL Transcripts for Auto QA Generation}
\begin{figure}[!t]
    \centering
    \includegraphics[width=0.7\columnwidth]{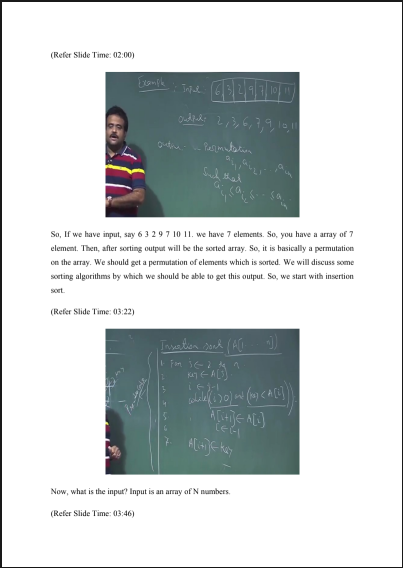}
    \caption{Example page from NPTEL transcript}
    \label{fig:nptel_transcript}
\end{figure}
The process of auto-generation of QA pairs from the video is a fairly complex process in practicality. To generate QA pairs, one would need to accurately sample frames from the video with maximum information and content, as well as make sure the transcripts are accurate and in sync with the video. To eliminate these problems, we choose a resource in NPTEL transcript documents, where this work has already been done by subject matter experts (SME) \footnote{\url{nptel.ac.in/aboutus}}, where the audio has been accurately transcribed, and key frames have been manually selected, and compiled into a single pdf file. We scrape these pdfs from the selected courses, convert the pages to images using online tools \footnote{\url{https://smallpdf.com/pdf-to-jpg}}. Referring to the example transcript in Figure \ref{fig:nptel_transcript}, we see the presence of transcribed text, key frames, along with associated timestamps. 

\subsubsection{Bloom's Taxonomy Definitions}
The Bloom's taxonomy definitions for auto-tagging using GPT4-o1-mini were inspired by past work \cite{forehand2010bloom}. We define them as follows: \\
\textit{1) Knowledge:} This initial level involves the simple recall of facts, terms, and basic concepts from memory. \\
\textit{2) Comprehension:} At this stage, understanding and grasping the meaning of information is key, including translating knowledge into one's own words. \\
\textit{3) Application:} This involves using learned material in new and concrete situations, applying rules, methods, concepts, principles, and theories. \\
\textit{4) Analysis:} This step is about examining and breaking information into component parts to understand its structure and relationships. \\
\textit{5) Synthesis:} This involves combining parts to form a new whole or propose alternative solutions. \\
\textit{6) Evaluation:} It involves making judgments about the value of ideas or materials based on criteria and standards through checking and critiquing. \\
While prompting the tagging model, we give the same definitions along with some example annotations which were made manually. The same instructions were given to the manual annotators.

\subsubsection{Adversarial Refinement Process}
The process of Adversarial Refinement of the generated questions is inspired by the process in \cite{rawal2024cinepile}. The goal was to extract questions which are difficult to answer without context. We follow the following process: \\
1) The QA pairs were fed to GPT-4o, without any additional frames or transcript content. \\
2) The answer generated A' were collected from the output. \\
3) For each answer, we calculate the Entailment Score(ES) \cite{ray2024ervqa} with respect to the originally generated answer, with a threshold above 0.50. \\
4) We pick 50 QAA' tuples from the generations, with the ES close to the threshold value for filtering (within 0.05). \\
5) We then compare the generated answer with the original answer, to see whether the general content of the answer matches - that is whether the original answer context matches with the generated answer context. \\
6) We perform this for threshold values of 0.5, 0.6, 0.7, 0.8 and 0.9. Since, for 0.6 about 20\% of the generated answers were capturing the context, and for 0.7, none of the contexts were matching, we decide to check with the threshold value of 0.65, where also none of the contexts were not matching. Since, we wanted to select a threshold that would yield the maximum number of valid QA pairs, we fixed the threshold to 0.65, and collected all the questions.

\subsection{Question Distributions}
\subsubsection{Distribution of Difficulty Tags}
\begin{figure}[!t]
    \centering
    \includegraphics[width=0.9\columnwidth]{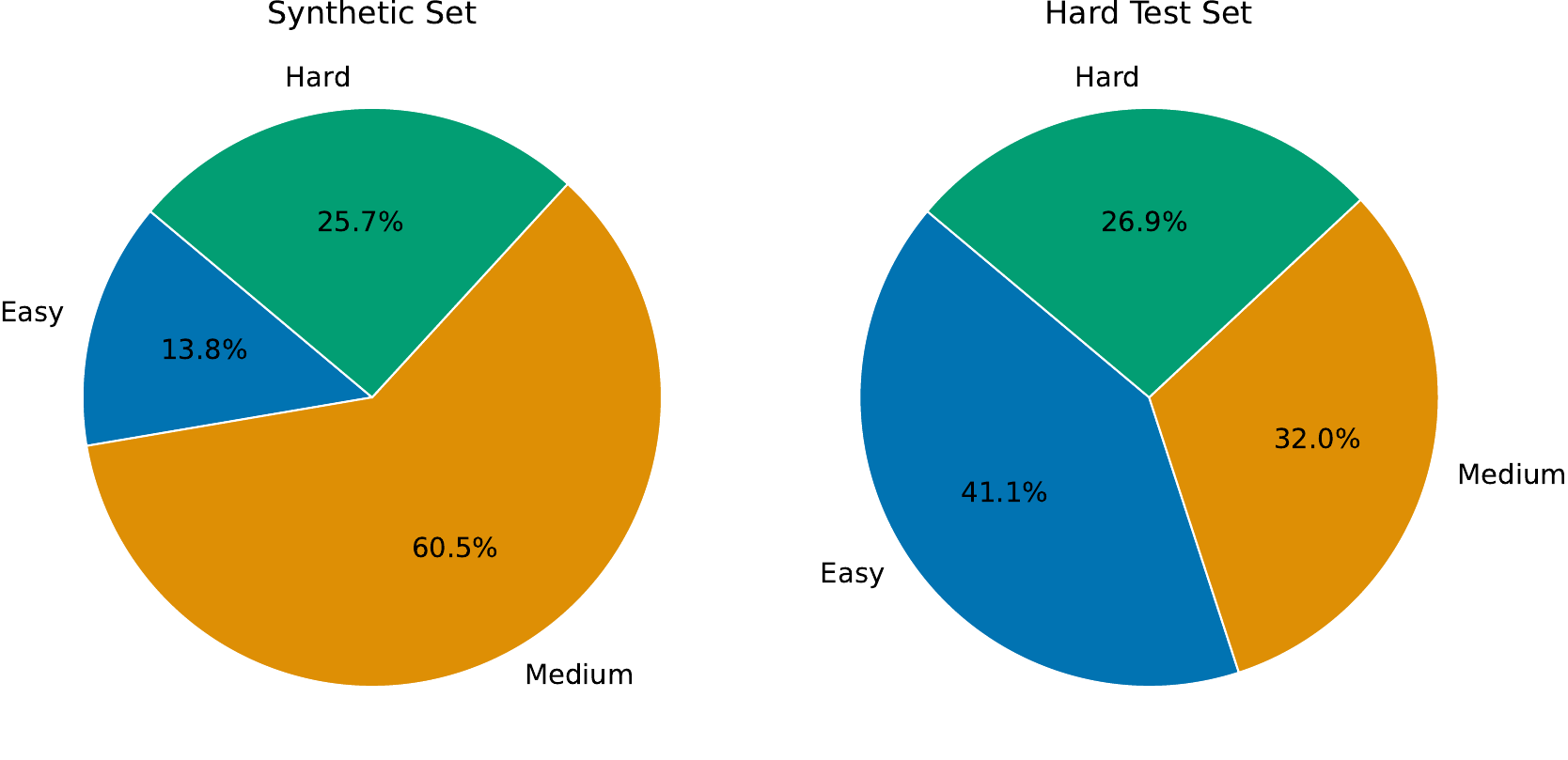}
    \caption{Question difficulty distribution in the Synthetic Set and the Real-World Test Set (referred to as 'hard')}
    \label{fig:difficulty}
\end{figure}
Figure \ref{fig:difficulty} shows the distribution of the question difficulty in the real-world set (referenced as 'Hard Test Set' in the figure). It shows that the real-world set has a more uniform distribution of difficulty tags, whereas the synthetic set has a large amount of 'medium' difficulty questions, compared to 'easy' and 'hard' questions. This makes the observations made through Study 2 (Appendix D.2) even more significant, as a large amount of questions needed to be edited to ensure greater quality ( as discussed in Section 5.2.2). 
\subsubsection{Qualitative Assessment of Questions}
While using the qualitative metrics described in Section 5.2.2, we observe the following points. The average scores for Clarity for the real-world test set is 3.58, and for the synthetic set is 4.70. This is difference is perhaps because the synthetic set has been explicitly modified to enhance clarity. This is desired as we would want clarity to be maximized in the answers through the supervised finetuning set. The average scores for Encouraging Critical Thinking (ECT) for the real-world set is 1.17 and synthetic set is 1.32. Similarly, average Using Pedagogical Techniques (UPT) scores are 1.19 for the real-world set and 1.53 for the synthetic set. These scores being relatively close is also desired, as it shows that the real-world set and the synthetic set are similar when it comes to these qualities, and training on the synthetic set can be useful.

\subsubsection{Discussion on Visual Dependence}
The authors intended to understand the visual dependence of questions in the synthetic set and real-world set. Manual checking revealed 44\% of the real world set contained textual queues indicating visual dependence, such as \textit{"According to the slide"}, \textit{"In the frame"}, and so on, and 46\% of the synthetic set showed the same. However, closer analysis of the video content also revealed that the context of many of these questions can also be gained from the textual transcript of the video. Moreover, models were also sometimes able to guess the context just from the question - a phenomenon that we wish to explore in our future work. Hence, while designing our experimental setups, we include both frames and transcripts, as the purpose of this work is not to focus on the spatio-temporal aspect of the task, but is that of being a stepping stone towards building utilitarian systems for students.   
\section{Details of Human Studies}
In this section, we describe some additional details and considerations from the Qualitative Preference Study (Section 4), along with a detailed description and results of a study done on the Relation between Answer Qualities and Question Difficulty.

\subsection{Additional Details of Qualitative Preference Study (from Section 4)}
\begin{table*}[!htbp]
\centering
\small
\renewcommand{\arraystretch}{1.5}
\begin{tabular}{|p{0.30\textwidth}|p{0.30\textwidth}|p{0.30\textwidth}|}
\hline
\textbf{Question} & \textbf{Answer (Option 1)} & \textbf{Answer (Option 2)} \\
\hline
In the video at 04:49, the presenter says, "nobody can come with some input where my code is performing worse than what I am claiming." In the context of algorithm analysis, why is it important to consider the worst-case scenario, and how does it benefit the developers and users of the algorithm? & The importance of considering the worst-case scenario, as the instructor states, lies in its ability to provide a bound on the algorithm's performance guarantee. For developers, it offers a conservative estimate that ensures the algorithm will not exceed the calculated maximum time under any circumstances. This is crucial for real-time systems or applications where performance guarantees are necessary. For users, understanding the worst-case helps set the right expectations and in the selection of the suitable algorithm for a particular use-case. By preparing for the worst-case, developers can ensure system stability and reliability, since they can be confident that the running time will not unpredictably exceed the worst-case estimate. & Considering the worst-case scenario is important because it helps developers know how bad the algorithm can get. It ensures that the algorithm won't take longer than expected in the worst situation. This is helpful for users because they can understand what to expect from the algorithm. Preparing for the worst-case makes sure the system works okay even in tough situations. \\
\hline
\end{tabular}
\caption{Example of choices given to students during the Qualitative Enhancement Study (Study 1). Students were asked to choose their preferred version of the answer.}
\label{tab:study_1_example}
\end{table*}
\begin{figure}[!t]
    \centering
    \includegraphics[width=0.9\columnwidth]{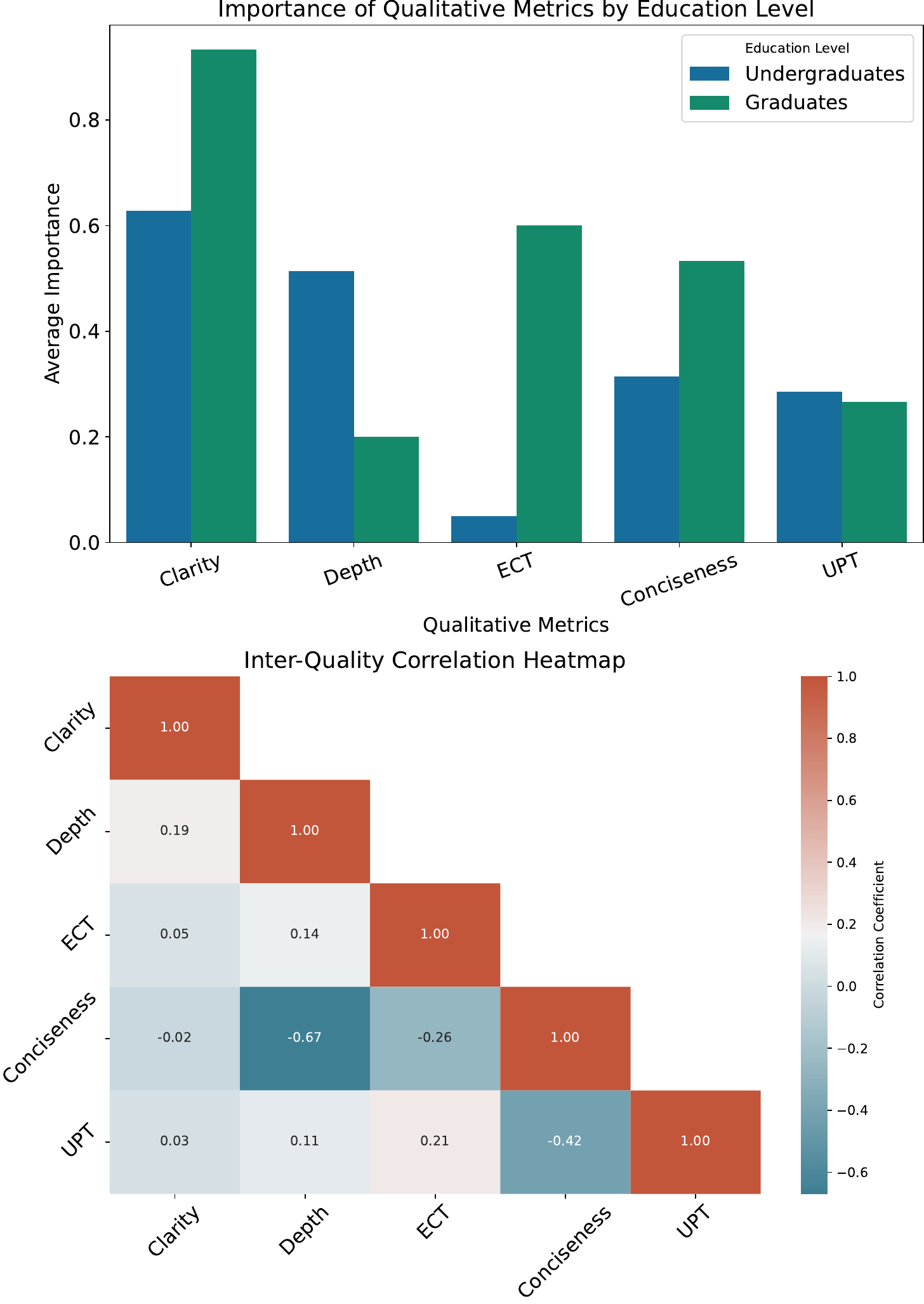}
    \caption{Results of Study 1 on the Importance of Different Quality Metrics. \textbf{\textit{Top: }}Shows the average preference per query for each quality for undergraduates and graduates. \textbf{\textit{Bottom:}} shows the Kendall's Tau correlation between the different quality metrics according to the student responses.  }
    \label{fig:study1_additional}
\end{figure}
With regards to the Qualitative Preference Study, we would like to show the following additional details: \\
\textbf{Additional Observations: } \\
\textbf{a) Graduates prefer Conciseness, Undergraduates prefer Depth:} In Figure \ref{fig:study1_additional} (Top), we see that there is a clear distinction between undergraduates and graduates, when evaluating preferences for Depth and Conciseness. Keep in mind, Depth and Conciseness are essentially opposite of each other, and we wished to see if there was a consensus in either one of them, which we could then utilize for our work. The Average Importance, calculated by the preference per data point, per annotator, shows that graduate students give more importance to Conciseness and undergraduate students give more importance to Depth. This is likely due to the difference in experience level in the subject matter along with their confidence in grasping new concepts. This is the reason, we do not consider these factors in our evaluation metrics and answer editing process, as we intend to capture the qualitative aspects that are universal to all students. To this regard, one might also note the difference in Importance with regards to Encouraging Critical Thinking. However, since graduate students give a high importance to the factors, and that it is a singular quality without any inverse, we consider it to be important for our subsequent work. \\
\textbf{b) Correlation between the Qualities: } Figure \ref{fig:study1_additional} (Bottom) shows the correlation between the different qualities. This is shown to confirm the validity of our definitions of each quality with regards to annotations. As expected, there is a strong negative correlation between Depth and Conciseness. Also, there is a moderate negative correlation between Conciseness, and Encouraging Critical Thinking and Using Pedagogical Techniques. This is also expected, as Critical Thinking and Pedagogical Thinking both affect the verbosity of the answer. We provide an example of the choices given to the students in Table \ref{tab:study_1_example}.

\subsection{Study 2: Relation between Answer Qualities and Question Difficulty}

\begin{figure}[!t]
    \centering
    \includegraphics[width=0.9\columnwidth]{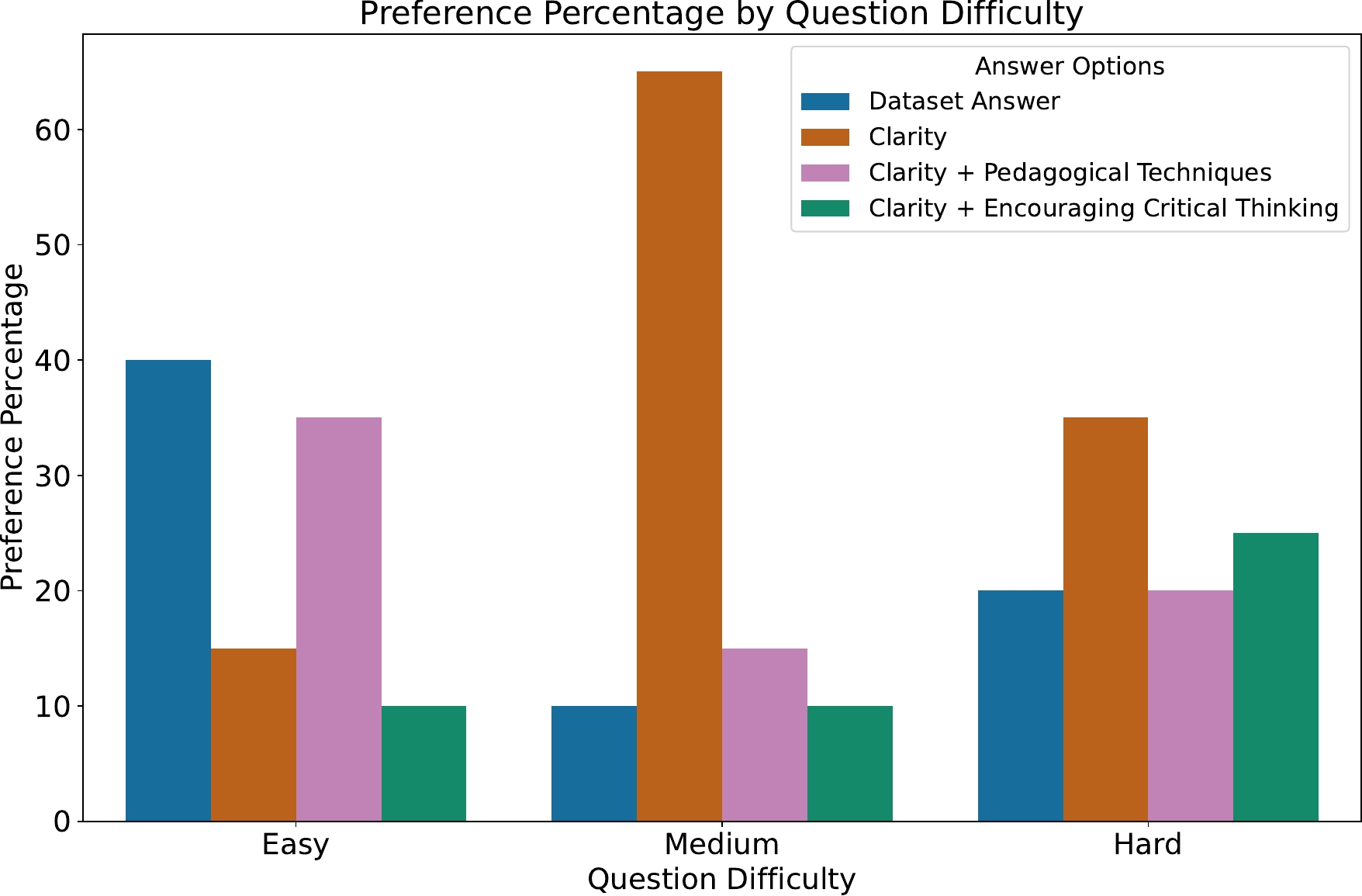}
    \caption{Results of Study 2 on relation between Answer Qualities and Question Difficulty}
    \label{fig:qstudy2}
\end{figure}
\textbf{Aim: }The aim of Study 2 was to analyze how enhancing specific qualitative features in answers—identified in the Qualitative Preference Study—impacts student preferences across different question difficulty levels (easy, medium, hard). This follow-up experiment was designed to empirically assess whether improving qualities like Clarity, Pedagogical Techniques, and Encouraging Critical Thinking makes answers more appealing to students. \\
\textbf{Setting:} We provided 20 students with three questions of varying difficulty (easy, medium, hard). Given Clarity was the most preferred quality, we enhanced it using GPT-4o. Since \textit{Uses Pedagogical Techniques} and \textit{Encourages Critical Thinking} are weakly correlated (see Figure \ref{fig:qstudy2}), we further enhanced these qualities separately. Students were given four answer options per question:  \\
(1) Original dataset answer  \\
(2) Clarity-enhanced answer (GPT-4o) \\  
(3) Clarity + Pedagogical Techniques (GPT-4o) \\  
(4) Clarity + Critical Thinking (GPT-4o)   \\
\textbf{Results:} For easy questions, 35\% preferred Option 3, while 40\% chose the original answer (Option 1). Medium questions saw a strong preference (65\%) for the clarity-enhanced answer (Option 2). For hard questions, 35\% preferred Option 2, but preferences were unclear overall for easy and hard questions. However, as hypothesized in Study 1, medium-level questions benefited from added clarity. Based on this, we enhance clarity for all medium-level questions in our synthetic dataset. This study, although relatively small scale, provides a valuable resource for future research. Examples data points from the study are given in Table \ref{tab:study_2_example}. \\
\textbf{Why not enhance all answers? } The reason we do not want to blindly enhance all answers using GPT4o is because 1) The originally generated answer might alreay possess these qualities to an appropriate amount. 2) Not all answers require an enhancement. For example for a question belonging to the Easy category (Q: What is the return type of len() function in python?), such enhancements might be overkill. We wanted to preserve these nuances in the dataset. 
\begin{table*}[!htbp]
\centering
\tiny
\renewcommand{\arraystretch}{1.5}
\begin{tabular}{|p{0.08\textwidth}|p{0.05\textwidth}|p{0.17\textwidth}|p{0.17\textwidth}|p{0.17\textwidth}|p{0.17\textwidth}|}
\hline
\textbf{Question} & \textbf{Difficulty level} & \textbf{Original Answer} & \textbf{Clarity enhanced} & \textbf{Clarity + ECT enhanced} & \textbf{Clarity + UPT enhanced} \\
\hline
At 05:20, can you explain how the Fixed Universe Successor problem differs from the dynamic set problems seen in binary search trees, given that both involve a set of operations like insert, delete, and find successor? & Easy & The primary difference between the Fixed Universe Successor problem and typical dynamic set operations in binary search trees lies in the nature of the universe from which elements are drawn. In the Fixed Universe Successor problem, all elements come from a pre-defined fixed universe, which is usually a consecutive range of integers. This constraint means that the universe is static and does not change over time. In contrast, with binary search trees, the elements can come from any arbitrary, potentially infinite, set without such restrictions. This allows for more flexibility but also introduces complexity concerning balancing and restructuring the tree after insertions and deletions to maintain efficient operation times. The fixed universe allows for some optimizations due to its predictability and finiteness that are not possible in a more general setting. & The main difference between the Fixed Universe Successor problem and typical dynamic set operations in binary search trees is the source of elements they use. In the Fixed Universe Successor problem, all elements come from a pre-defined fixed universe, usually a range of consecutive integers. This means that the universe remains static and unchanging over time. On the other hand, binary search trees can draw elements from any infinite, arbitrary set, leading to increased flexibility but also complexity when balancing and restructuring the tree after operations like insertions and deletions to maintain efficient performance. While the fixed universe allows for certain optimizations due to its predictability and limited size that are not possible in a more general setting. & To improve the critical thinking aspect of this answer, we can consider alternative perspectives and explore potential implications of using a fixed universe versus an arbitrary set. For example, using a fixed universe may simplify certain operations and allow for faster execution times, but it also limits the range of elements that can be considered, which might not always be ideal depending on the problem at hand. On the other hand, using an arbitrary set provides greater flexibility in selecting elements, but it can introduce complexity in managing the tree's structure and performance. It is essential to weigh these trade-offs and choose the appropriate approach based on the specific requirements and constraints of the problem being solved. & To enhance understanding, let's use an example. Imagine we have a binary search tree with elements ranging from 1 to 100. In this case, the dynamic set operations can easily be performed as the elements and their relationships are not predetermined. However, if we restrict the range of elements in our universe to a fixed range, such as integers between 1 and 100, the tree's structure becomes rigid, limiting the ability to optimize performance through dynamic set operations. This is the main difference between the Fixed Universe Successor problem and typical dynamic set operations in binary search trees. \\
\hline
At 15:30, the instructor mentioned that insertion in a sorted array has a time complexity of O(n). Can you explain why it isn't O(log n) since we can locate the position using binary search? & Medium & Although it’s true that binary search allows us to locate the position where the new element should be inserted in O(log n) time for a sorted array, the key aspect contributing to the time complexity here is the shifting of elements after the insert position. When we insert the new element, we need to shift all subsequent elements one position to the right to make space for the new element. In the worst-case scenario, this could mean shifting all n elements if the new element is to be inserted at the beginning of the array. Hence, the overall time complexity for insertion in a sorted array would be dominated by the shifting operation, which is O(n). & Binary search allows us to locate the position where a new element should be inserted into a sorted array in O(log n) time. However, the key aspect that contributes to the time complexity here is the shifting of elements after the insertion position. When inserting a new element, we need to shift all subsequent elements one position to the right to make space for the new element. In the worst-case scenario, this could mean shifting all n elements if the new element is to be inserted at the beginning of the array. Hence, the overall time complexity for insertion in a sorted array would be dominated by the shifting operation, which has a time complexity of O(n). & Binary search allows us to efficiently locate the position where a new element should be inserted into a sorted array in O(log n) time. However, it's crucial to recognize that the key factor contributing to this time complexity is the shifting of elements after the insertion position. When inserting a new element, we must shift all subsequent elements one position to the right to make space for the new element. In the worst-case scenario, this could involve shifting all n elements if the new element is to be inserted at the beginning of the array. Consequently, the overall time complexity for insertion in a sorted array would be primarily determined by the shifting operation, which has a time complexity of O(n). To improve critical thinking skills, it's important to consider alternative approaches and explore open-ended suggestions, such as using a different data structure or implementing a custom algorithm tailored to specific insertion scenarios. & To enhance understanding, let's break down the process of inserting an element into a sorted array using binary search. Firstly, we identify the position where the new element should be inserted. Then, we shift all subsequent elements one position to the right to make space for the new element. In the worst-case scenario, this could mean shifting all n elements if the new element is to be inserted at the beginning of the array. The time complexity of shifting elements is O(n), which dominates the overall time complexity for insertion in a sorted array. By using examples and step-by-step explanations, we can understand how binary search works and its efficiency in inserting elements into a sorted array in O(log n) time. \\
\hline
Q3: At 19:20, when discussing linked lists, the instructor states the time complexity for insert and delete operations is O(1). Can it always be assumed that insertions and deletions in a linked list are constant time, regardless of the position? & Hard & Insertions and deletions in a linked list have an average time complexity of O(1) when you are inserting or deleting elements at the head or directly at a known position with a reference (e.g., after a given node). This is because there's no need to shift the elements like in an array; you only update the pointers. However, if you need to insert or delete an element at a specific position and you do not have a reference to the node after or before which you want to perform the operation, you will first have to traverse the list to find the correct spot, which could take O(n) time in the worst case. Therefore, while the actual insertion or deletion is O(1), finding the position in the linked list where the operation needs to be performed may increase the overall time complexity. & Insertions and deletions in a linked list have an average time complexity of O(1) when done at the head or a known position with the help of a reference. This is because, unlike arrays, you only need to update pointers without shifting elements. However, if the insertion or deletion is done at a specific position and there's no reference for the target node, traversing the list to find the correct spot may take O(n) time in the worst case, making the overall complexity worse. & To improve the critical thinking aspect of this answer, we can discuss alternative approaches and suggest open-ended suggestions. One approach could be using a doubly linked list, where each node contains both a reference to the previous node and a reference to the next node. This allows for efficient insertions and deletions at any position without the need for traversing the entire list. Another approach is to use a hash table or binary search tree, which can provide constant time access to specific positions in the list. However, these data structures may have higher overhead in terms of memory usage or complexity. It's important to consider the trade-offs and choose the appropriate data structure based on the specific requirements and constraints of the application. & To enhance understanding, let's use an example to illustrate the concept of insertions and deletions in a linked list. Imagine we have a linked list containing the numbers 1 through 5, with each number pointing to the next node in the list. Now, let's say we want to insert a new node at index 3. With the help of a reference, we can easily insert the new node at index 3 without affecting the other nodes in the list. However, if we don't have a reference and need to find the correct position to insert the new node, we would need to traverse the entire linked list, which could take O(n) time in the worst case, making the overall complexity worse. Therefore, it's important to use references when inserting or deleting nodes at specific positions in a linked list to avoid worst-case scenarios and maintain efficient time complexity. \\
\hline
\end{tabular}
\caption{Examples of answer enhancements for varying difficulty levels, using clarity, explanation-through-contextual-thinking (ECT), and understanding pedagogical technique (UPT) strategies. Students were asked to choose their preferred version of the answer.}
\label{tab:study_2_example}
\end{table*}
\section{Benchmarking Details}

\subsection{Experiment Prompts}


We use the same experiment prompt across all models, during both supervised finetuning and zero-shot experiments. We define the prompt as follows: \\
\textbf{VLM Prompt:} \\
SYSTEM\_PROMPT = \textit{“You are an expert computer science educator. You have to answer a question that a student has asked from a video. For context, we have provided you with the transcript around the relevant timestamp, and the frame from the video corresponding to the relevant timestamp.”} \\
QUESTION\_PROMPT = \textit{“Make sure the answer has good clarity, uses pedagogical techniques and encourages critical thinking. Use the context from the transcript to answer the following question in a single paragraph. ”} \\
 final\_prompt = \textit{"System Prompt: "} + SYSTEM\_PROMPT  + \textit{" Relevant transcript: "} + transcript\_text  + \textit{"Question Prompt: "} + QUESTION\_PROMPT + \textit{"Question: "} + df['question'][i] \\
 \textbf{Video LLM Prompt:} \\
SYSTEM\_PROMPT = \textit{“You are an expert computer science educator. You have to answer a question that a student has asked from a video. For context, we have provided you with the transcript around the relevant timestamp, and the frames from the video corresponding to the relevant timestamp.”} \\
QUESTION\_PROMPT = \textit{“Make sure the answer has good clarity, uses pedagogical techniques and encourages critical thinking. Use the context from the transcript to answer the following question in a single paragraph. ”} \\
 final\_prompt = \textit{"System Prompt: "} + SYSTEM\_PROMPT  + \textit{" Relevant transcript: "} + transcript\_text  + \textit{"Question Prompt: "} + QUESTION\_PROMPT + \textit{"Question: "} + df['question'][i] 

\subsection{Evaluation Metric Details}

\subsubsection{FactQA metric Prompt}
We take the FactQA metric prompt from the SyllabusQA paper \cite{fernandez2024syllabusqa}, and slightly modify it for our use case. \\
\textbf{Prompt: } \\
\textit{
Your job is to evaluate the similarity of different answers to a single question. 
You will be given a question from a specific computer science college course. 
You will also be given two possible answers to that question, 
and will have to evaluate the claims in one answer against the other. \\
Steps: \\
1. List all of the atomic claims made by Answer 1. 
Note that an answer saying that there is no information counts as a single claim. \\
2. Tell me which of those claims are supported by Answer 2. \\
3. Summarize the results using the template: Score: <num supported claims>/<num total claims>
Ensure that both numbers are integers. \\
Question:} <question> \\
\textit{Answer 1:} <answer\_1> \\
\textit{Answer 2:} <answer\_2> 

\subsubsection{Qualitative Metric Details}
\textbf{a) Qualitative Metric Prompts: } \\
For annotating the model outputs using GPT-4, we use in-context example prompting. For each prompt, we define the objective, the Likert Scale, output format, in-context examples for each score on the scale and finally a query prompt. We generate a JSON object which we then parse to get the score. Following are the prompts: \\
\textbf{i) Clarity prompt:} \\
SYSTEM\_PROMPT = \textit{"You are a domain expert in Computer Science. You are given a question and an answer. Judge the answer and give an appropriate score."} \\
TASK\_INSTRUCTION = \textit{
"Task Instructions: You are a strict grader. Judge the answer on the following criterion:  
Clarity (simplifies complex terms, logically structured, unambiguous)  \\
Rules: \\
Count unexplained jargon terms -> Count transition phrases (e.g., "so", "therefore", "next") → Detect ambiguous or compound run-on statements. \\
Assign a score based on the following scale: \\
1 = >=2 jargon terms without explanation, and >=2 incoherent transitions. \\
2 = >= 1 jargon term unexplained and at least 1 logical jump or ambiguous phrasing. \\
3 = Mostly clear, but 1–2 minor issues: one ambiguous phrase or slightly choppy flow. \\
4 = All terms explained, clear flow, no ambiguity except possibly 1 unclear phrase. \\
5 = No unexplained jargon, consistent logical flow, zero ambiguity."
} \\
OUTPUT\_FORMAT = \textit{
Output Format: Return **only** the JSON object below—no extra text. \\
\{
"explanation": <explanation>,
  "overall": <1-5>
\}
} \\
examples = \textit{ '''
Here are some examples:
Example 1:
Input: [Q]: Explain what a Binary Search Tree (BST) is and why it is useful? Answer: BSTs optimise O(log n) retrieval; apex node bifurcates sub-trees ergo bigger left subchild contrarily right.
Output: \{
"explanation": "Two jargon terms (O(log n), apex), two incoherent jumps (“ergo”, “contrarily”): fails both thresholds.",
"overall": 1
\} \\
Example 2:
Input: [Q]: Explain what a Binary Search Tree (BST) is and why it is useful? Answer: A BST has nodes and children. Therefore, data is stored efficiently.
Output: \{
"explanation": "One unexplained term (nodes), one logical jump (why does it imply efficiency?)",
"overall": 2
\} \\
Example 3:
Input: [Q]: Explain what a Binary Search Tree (BST) is and why it is useful? Answer: A BST is a tree where each left child holds a smaller value than its parent, and the right child a larger value. This rule lets us skip half the tree each step, making searches fast.
Output: \{
"explanation": "Mostly clear, but phrase “skip half the tree” is mildly ambiguous about how.",
"overall": 3
\} \\
Example 4:
Input: [Q]: Explain what a Binary Search Tree (BST) is and why it is useful? Answer: A Binary Search Tree (BST) is an ordered tree: every node’s left subtree contains only smaller keys, the right subtree only larger keys. Following this rule top-down lets you discard half the remaining elements each comparison.
Output: \{
"explanation": "All terms defined; flow is clear; one sentence is long but unambiguous.",
"overall": 4
\} } \\
Example 5:
Input: [Q]: Explain what a Binary Search Tree (BST) is and why it is useful? Answer: A Binary Search Tree (BST) is a sorted tree structure. For each node: left <parent<right. Starting at the root you compare the target key: if it is smaller, move left; if larger, move right. Repeating this till you reach a leaf takes at most log n steps. No jargon remains unexplained, and each step follows directly from the previous one.
Output: \{
"explanation": "Zero jargon, crisp stepwise flow, no ambiguity.",
"overall": 5
\}   
 \\
QUERY\_PROMPT = \textit{"Given the following question and answers, assign the appropriate score and give the explanation as shown in the examples. Output the dictionary format as described above. Do not include any other text. "} \\
final\_prompt = SYSTEM\_PROMPT + TASK\_INSTRUCTION + OUTPUT\_FORMAT + examples + QUERY\_PROMPT + \textit{"[Q]: "} + df['question'][i] + \textit{"Answer: "} + df['generated'][i] + \textit{" Output: "} \\ 
\textbf{ii) ECT Prompt: } \\
SYSTEM\_PROMPT = \textit{"You are a domain expert in Computer Science. You are given a question and an answer. Judge the answer and give an appropriate score."} \\
TASK\_INSTRUCTION = \textit{
Task Instructions: You are a strict grader. Judge the answer on the following criterion:  
Encouraging Critical Thinking (Prompts learners to reflect, explore alternatives, or ask new questions.) \\
Rules: \\
Detect open-ended question marks ("why", "how", "what if"), alternatives (“another way”, “alternatively”, “one approach is…”), and exploration prompts (“you may explore”, “consider trying…”). \\
Assign a score based on the following scale: \\
1 = No questions, no alternatives, purely factual. \\  
2 = Includes 1 suggestive or reflective phrase, but no actual open-ended question. \\
3 = Contains 1 open-ended question or 1 alternative method/viewpoint. \\
4 = >=2 open-ended prompts or multiple viewpoints briefly compared. \\
5 = >=2 open-ended questions + explicit invitation to explore further. \\
} \\
OUTPUT\_FORMAT = \textit{
Output Format: Return **only** the JSON object below—no extra text. \\
\{
"explanation": <explanation>,
  "overall": <1-5>
\}
} \\
examples = \textit{ '''
Here are some examples:
Example 1:
Input: [Q]: Input: [Q]: Explain what a Binary Search Tree (BST) is and why it is useful? Answer: BSTs are efficient search structures used in programming.
Output: \{
"explanation": "Purely factual; no questions or alternatives.",
"overall": 1
\} \\
Example 2:
Input: Explain what a Binary Search Tree (BST) is and why it is useful? Answer: BSTs are efficient; it is worth thinking about their balance.
Output: \{
"explanation": "Reflective phrase “worth thinking” but no open-ended question.",
"overall": 2
\} \\
Example 3:
Input: [Q]: Explain what a Binary Search Tree (BST) is and why it is useful? Answer: How would search time change if the tree became unbalanced?
Output: \{
"explanation": "One open-ended question prompts reflection.",
"overall": 3
\} \\
Example 4:
Input: [Q]: Explain what a Binary Search Tree (BST) is and why it is useful? Answer: What happens if the tree degenerates into a list—and can you think of another structure that avoids this? Compare that with self‑balancing trees such as AVL.
Output: \{
"explanation": "Two prompts: a “what happens” question plus an alternative to explore.",
"overall": 4
\} \\
Example 5:
Input: [Q]: Explain what a Binary Search Tree (BST) is and why it is useful? Answer: Why might a hash table outperform a BST for large data sets? After trying a BST yourself, consider experimenting with AVL or Red‑Black trees and evaluate which conditions favour each structure.
Output: \{
"explanation": ">=2 open questions and an explicit invitation to explore further.",
"overall": 5
\}"  
} \\
QUERY\_PROMPT = \textit{"Given the following question and answers, assign the appropriate score and give the explanation as shown in the examples. Output the dictionary format as described above. Do not include any other text. "} \\
final\_prompt = SYSTEM\_PROMPT + TASK\_INSTRUCTION + OUTPUT\_FORMAT + examples + QUERY\_PROMPT + \textit{"[Q]: "} + df['question'][i] + \textit{"Answer: "} + df['generated'][i] + \textit{" Output: "} \\
\textbf{iii) UPT Prompt: } \\
SYSTEM\_PROMPT = \textit{"You are a domain expert in Computer Science. You are given a question and an answer. Judge the answer and give an appropriate score."} \\
TASK\_INSTRUCTION = \textit{
Task Instructions: You are a strict grader. Judge the answer on the following criterion:  
Using Pedagogical Techniques (Employs examples, analogies, or step-by-step explanations to aid understanding) \\
Rules: \\
Search for example phrases (“for example”, “e.g.”), analogies (“like”, “similar to”), step phrases (“Step 1”, “First,” “Then”).
 \\
Assign a score based on the following scale: \\
1 = Pure explanation without any example or breakdown. \\
2 = 1 brief example or partial list of steps, lacking clarity. \\
3 = 1 complete example or full step list present, but not both. \\
4 = >=2 teaching techniques used (e.g., example + step list), with moderate clarity. \\
5 = >=3 techniques (e.g., example, analogy, visual mention), all clear and complete. \\
} \\
OUTPUT\_FORMAT = \textit{
Output Format: Return **only** the JSON object below—no extra text. \\
\{
"explanation": <explanation>,
  "overall": <1-5>
\}
} \\
examples = \textit{ '''
Here are some examples:
Example 1:
Input: [Q]: Explain what a Binary Search Tree (BST) is and why it is useful? Answer: BSTs let you perform searches in O(log n) time.
Output: \{
"explanation": "No example, no steps, no analogy.",
"overall": 1
\} \\
Example 2:
Input: [Q]: Explain what a Binary Search Tree (BST) is and why it is useful? Answer: BSTs let you search, e.g., finding a student ID quickly.
Output: \{
"explanation": "One brief example only.",
"overall": 2
\} \\
Example 3:
Input: [Q]: Explain what a Binary Search Tree (BST) is and why it is useful? Answer: Step 1: Start at root. Step 2: Compare key. Step 3: Move left or right until found.
Output: \{
"explanation": "Full step list but no example/analogy.",
"overall": 3
\} \\
Example 4:
Input: [Q]: Explain what a Binary Search Tree (BST) is and why it is useful? Answer: Imagine a phone book sorted alphabetically (analogy). Step 1… Step 3… Finally, for example, you can locate roll-number 73 in ~7 comparisons.
Output: \{
"explanation": "Two devices (analogy + steps) with good clarity.",
"overall": 4
\} \\
Example 5:
Input: [Q]: Explain what a Binary Search Tree (BST) is and why it is useful? Answer: “Think of a BST like a decision fork in a 20‑questions game (analogy). Example: searching 42 follows arrows in the diagram below. Steps: 1 Start at 50, 2 go left… Visual mention: 'See diagram'.
Output: \{
"explanation": "Analogy + example + step list (+ visual cue) three devices, all complete.",
"overall": 5
\}"  
} \\
QUERY\_PROMPT = \textit{"Given the following question and answers, assign the appropriate score and give the explanation as shown in the examples. Output the dictionary format as described above. Do not include any other text. "} \\
final\_prompt = SYSTEM\_PROMPT + TASK\_INSTRUCTION + OUTPUT\_FORMAT + examples + QUERY\_PROMPT + \textit{"[Q]: "} + df['question'][i] + \textit{"Answer: "} + df['generated'][i] + \textit{" Output: "} \\
\textbf{b) Instructions for Human Annotators:} \\
We essentially provide the same information to the annotators, as we provide in the prompts - the Likert Scale, Examples and some guidelines. Additionally, we define a pipeline to help them annotate the questions accurately. \\
\textbf{Annotation Pipeline: } \\
1. Read the entire answer once without scoring. \\
2. Highlight jargon (domain‑specific terms). Mark whether each is explained in‑line. \\
3. Highlight transitions and check logical flow. \\
4. Mark question marks and phrases signalling reflection or alternatives. \\
5. Tag teaching devices (examples, analogies, step lists, visual references). \\
6. Apply the decision tables provided to pick scores. Edge‑case rule: if an answer sits exactly on the boundary, choose the lower score to stay conservative. \\
7. Record the scores. \\
\textbf{Definitions of Ambiguous Terms: } \\
There are some terms in the definition of the Likert Scale in Section 5.2. We formally define them in order to avoid any confusion between the annotators: \\
\textbf{1. Jargon:} Jargon terms are domain-specific technical terms that may not be immediately understandable to a general audience or novice learners unless they are explicitly explained. If a term is used without explanation and it is not commonly understood outside a specific field, it counts as unexplained jargon. \\
\textit{\textbf{Example:} In "BSTs optimize O(log n) retrieval.", BSTs and O(log n) can be considered as jargon terms.} \\
\textbf{2. Logical Flow:} Logical flow refers to the coherent and sequential structure of ideas, where each sentence or step follows naturally from the previous one. It is disrupted by abrupt transitions or disconnected reasoning. Check if transitional phrases like “so,” “therefore,” or “next” logically connect to the prior statement. Count jumps or breaks in reasoning. \\
\textit{\textbf{Example of Poor Logical Flow:} “A BST has nodes and children. Therefore, data is stored efficiently.” - The "therefore" does not follow logically - why does having nodes imply efficiency?} \\
\textit{\textbf{Example of Good Logical Flow:} “For each node: left<parent<right. Starting at the root you compare the target key...” - Each step naturally builds on the last.} \\
\textbf{3. Suggestive Phrase:} A suggestive phrase hints at reflection or further exploration without forming a direct open-ended question. It nudges the learner toward deeper thought but does not demand a specific response. \\
\textit{\textbf{Example:} “It is worth thinking about their balance.” - Encourages reflection without asking a direct question.} \\
\textbf{4. Reflective Phrase:} A reflective phrase invites the learner to think back on a concept or consider implications, usually in a subtle or implicit way. It overlaps with suggestive phrases but emphasizes inward thought. \\
\textit{\textbf{Example:} “BSTs are efficient; it is worth thinking about their balance.” - The phrase invites internal consideration of balance.} \\
\textbf{5. Open-ended Question:} An open-ended question is one that cannot be answered with a simple yes/no or fact, and instead prompts exploration, reasoning, or analysis. It often starts with “how,” “why,” or “what if.” Look for phrases like: “How does…?”, “Why would…?”, “What happens if…?” \\
\textit{\textbf{Example:} “How would search time change if the tree became unbalanced?” - Encourages reflection and understanding beyond the surface.} \\
\textbf{\textit{Please note that these definitions might differ slightly from textbook definitions of these terms. We carefully design these, tailored to our unique use case.}} 

\subsubsection{Alternate Method for Qualitative Metrics}
\begin{table*}[!htbp]
\centering
\tiny
\renewcommand{\arraystretch}{1.5}
\setlength{\tabcolsep}{0.5em}
\begin{tabular}{p{0.05\textwidth}p{0.30\textwidth}p{0.30\textwidth}p{0.30\textwidth}}
\toprule
\textbf{Score} & \textbf{Clarity Scale} & \textbf{ECT Scale} & \textbf{UPT Scale} \\
\midrule
1 & Uses jargon without explanation; ideas jump randomly; meaning often unclear & Merely states facts; no prompts, no mention of alternatives; discourages inquiry. & Pure exposition; no examples or stepwise guidance.\\

2 & Many unexplained terms; some logical gaps; frequent ambiguous phrasing. & Sporadic “food-for-thought” phrases, but mostly didactic; few or no open questions. & One short example or a partial step list, but insufficient detail or relevance. \\

3 & Overall flow understandable though occasionally choppy; a few unclear parts. & Provides at least one open-ended question or discusses one alternative viewpoint. & Includes at least one concrete example and outlines main steps, though either could be clearer. \\

4 & Almost all terms explained in plain language; ideas build logically from point to point; wording rarely ambiguous. & Regularly poses thoughtful questions and briefly compares multiple perspectives or solutions. & Multiple relevant examples or a clear, complete step-by-step breakdown; occasionally uses analogies. \\

5 & Explains every term in learner-friendly language and shows how concepts fit together; transitions are seamless; no ambiguity detected. & Integrates probing questions throughout; systematically explores several alternatives; explicitly invites learners to investigate further. & Combines varied, well-chosen examples, analogies, visuals (if applicable), and a thorough step-by-step scaffold that anticipates learner difficulties. \\
\bottomrule
\end{tabular}
\caption{Likert Scales for Clarity, ECT and UPT, with more subjective definitions.}
\label{tab:likert_scale_2}
\end{table*}
\begin{figure*}[!t]
    \centering
    \includegraphics[width=\textwidth]{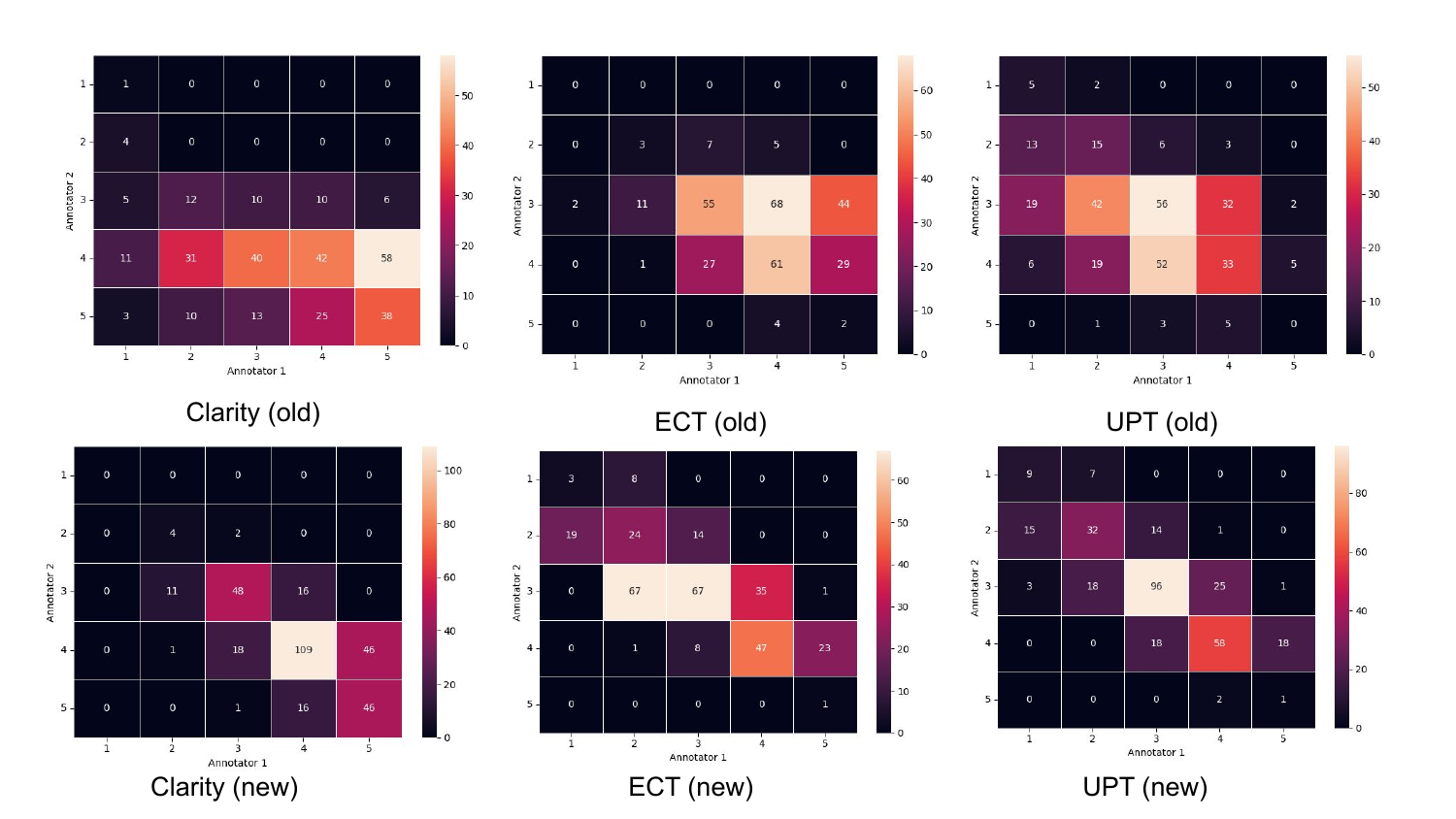}
    \caption{The difference in confusion matrices for subjective and more objective definitions of the qualitative metrics. Here, \textit{old} means the subjective definition scale and \textit{new} means the more objective scale used in our work.  }
    \label{fig:study1_additional}
\end{figure*}
In order to truly understand the effectiveness of our Likert Scale, we experiment with another Likert Scale where the definitions of the scores are more subjective. We show this Likert Scale in Figure \ref{tab:likert_scale_2}. In this case as well, we get two student annotators to annotate according to the scale, and check the results. \\
\textbf{Reasons for Rejection: } We observe that the annotations between the 2 annotators differ greatly, and this is rectified in the more objective version of the Likert Scale. For clarity, the Spearman's $\rho$ was 0.3654, for ECT 0.3878 and for UPT, 0.3764. Hence, we use the more objective scale for our study, and to design the evaluation metrics. 
\section{Additional Analyses}
\subsection{Difficulty-based Analysis}
\begin{figure}[!t]
    \centering
    \includegraphics[width=\columnwidth]{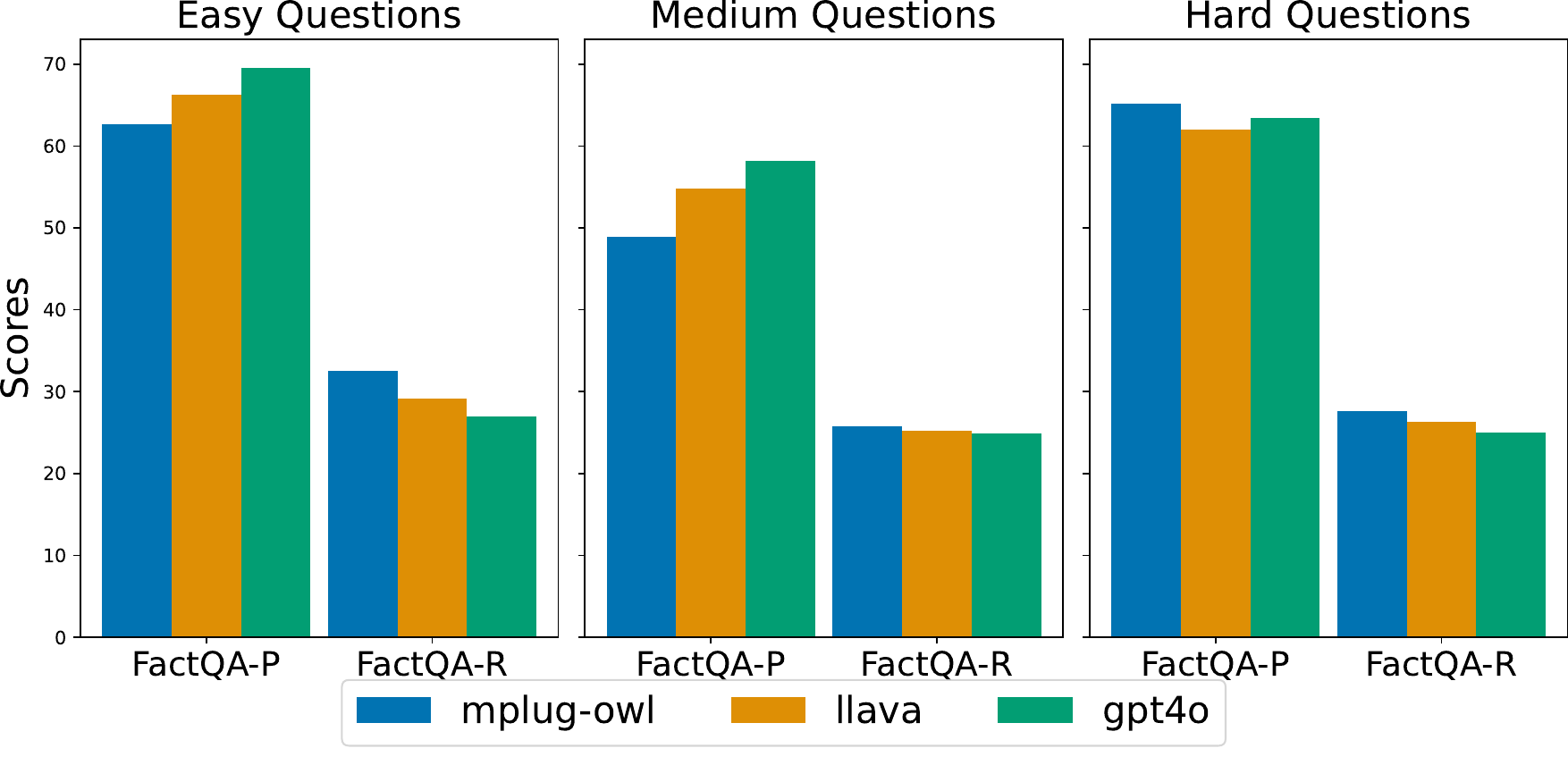}
    \caption{Performance of the best models from each model category against different question difficulty levels in the real world test set }
    \label{fig:difficulty_analysis}    
\end{figure}
Based on the categorization of question difficulty explained in Section 3.2, we analyze the performance of the top performing open source LVLM Llava-13B, closed LVLM GPT4o and open source Video LLM mPLUG-Owl on the real world test set, on the basis of FactQA-Precision and FactQA-Recall. The results are presented in Figure \ref{fig:difficulty_analysis}. Here, we observe an interesting result: all 3 models perform better on \textit{hard} questions compared to \textit{medium} questions, in terms of FactQA-Precision. This means that for the \textit{medium} and \textit{hard} questions, the answers are factually correct, without too much unnecessary text, as compared to \textit{medium} questions.  FactQA-Recall performance is relatively similar across all difficulty levels. This indicates these models are not able to get all the facts from the reference answer, regardless of the difficulty level. Nevertheless, an interesting point to note is that for \textit{hard} questions, mPLUG-Owl outperforms GPT4o across both metrics, which holds great significance in our attempts to use synthetic data for training.

\subsection{Reverse Question Answering Analysis}
\begin{table}[!htbp]
\centering
\small
\renewcommand{\arraystretch}{1.5}
\begin{tabular}{lc}
\toprule
\textbf{Model Name} & \textbf{Entailment Score (RQA)} \\
\midrule
GPT-4o & 0.2834 \\
mPLUG Owl-0shot & 0.2336 \\
mPLUG Owl-SFT & 0.1926 \\
Llava-0shot & 0.2178 \\
Llava-SFT & 0.2032 \\
\bottomrule
\end{tabular}
\caption{Reverse QA analysis}
\label{tab:rqa}
\end{table}

Building on the research objective shown in \cite{balepur2024reverse}, we use Reverse Question Answering as a potential evaluation objective of our models. This is some preliminary work that is meant to give some insights into the finetuning process, Further studies can be done to analyse the effectiveness of this method. \\
\textbf{Evaluation Methodology: } We provide the answer A without the question Q to GPT-4o, and ask it to predict the question. It gives a reverse engineered question Q', which we compare with Q, using Entailment Score (ES) \cite{ray2024ervqa}. We do this process 5 times for each answer, and take the Q' with the highest ES for our evaluation, in order to understand whether the \textit{answer A truly corresponds to the question Q}. \\
\textbf{Results: } Table \ref{tab:rqa} shows the results for this study. Out of all the best performing models from the benchmark, we see that GPT-4o performs best. Another interesting observation is that on finetuning the models, the ES goes down, possibly due to the answers becoming more verbose, in order to incorporate the qualitative aspects introduced through the synthetic set. \\
\textbf{Implication: } This study somewhat implies that the models, whether closed-source or open-source, possibly misunderstands the objective of the question to a certain degree, and hence prompts further investigation into the matter.  
\end{document}